\documentclass[10pt,twocolumn,letterpaper]{article}
\usepackage[T1]{fontenc}
\usepackage{lmodern}
\usepackage[margin=0.75in,columnsep=0.25in]{geometry}
\usepackage{graphicx}
\usepackage{amsmath,amssymb}
\usepackage{array}
\usepackage{booktabs}
\usepackage[font=small,labelfont=bf]{caption}
\usepackage{placeins}
\usepackage{microtype}
\usepackage[numbers,sort&compress]{natbib}
\usepackage{xcolor}
\definecolor{linkblue}{HTML}{174F86}
\usepackage[colorlinks,allcolors=linkblue]{hyperref}

\newcommand{\ours}{{\upshape\textsc{SlotNarrative}}}
\newcommand{\tok}{\#Tok}

\title{SlotNarrative: Persistent Object Narratives\\for Video-Language Understanding}
\author{%
  Junzhe Chen\textsuperscript{*}\qquad Siyuan Meng\qquad Xiaojie Guo\textsuperscript{\textdagger}\\[4pt]
  Tianjin University\\[2pt]
  \small\textsuperscript{*}\href{mailto:junzhec@tju.edu.cn}{\nolinkurl{junzhec@tju.edu.cn}}
  \qquad
  \small\textsuperscript{\textdagger}\href{mailto:xj.max.guo@gmail.com}{\nolinkurl{xj.max.guo@gmail.com}}%
}
\date{}

\hypersetup{
  pdftitle={SlotNarrative: Persistent Object Narratives for Video-Language Understanding},
  pdfauthor={Junzhe Chen, Siyuan Meng, Xiaojie Guo}
}

\begin{document}
\raggedbottom
\maketitle
\begin{abstract}
Video large language models (Video-LLMs) often receive videos as sequences of frame or patch features. These inputs retain visual evidence but leave implicit which observations belong to the same object and how its state changes over time. We introduce \ours{}, an object-centric Video-LLM that represents a video as \emph{persistent object narratives}. Recurrent slot attention groups frame features into object-like observations, and a training-free memory links them across time into clip-level object narratives. Each narrative enters the language model as one \emph{object identity token}, a clip-level appearance reference, and several \emph{object state tokens} that describe its appearance, geometry, and motion within temporal segments. Shared object embeddings bind the two token types, and segment embeddings locate each state in time. The interface is aligned with a frozen language model on video instruction data. Its length depends on object capacity and segment count, not on the number of frames, which keeps the cached language-model context small. With only 144 visual tokens, \ours{} achieves 75.6\%, 69.8\%, and 55.2\% accuracy on MSVD-QA, MSRVTT-QA, and ActivityNet-QA. Average pooling with the same backbones, data, and budget is 6.9 points lower on ActivityNet-QA. Controlled ablations show that memory improves accuracy at the same token budget and that identity and state tokens provide complementary evidence.

\end{abstract}

\smallskip
\noindent\textbf{Code:} \href{https://github.com/junzhe-c/slotnarrative}{\nolinkurl{github.com/junzhe-c/slotnarrative}}\par

\section{Introduction}
\label{sec:introduction}

Video large language models (Video-LLMs) extend pretrained large language models (LLMs) to open-ended video understanding by converting visual observations into language-compatible tokens. Representative systems, including Video-ChatGPT, Video-LLaVA, VideoLLaMA~2, and LLaVA-OneVision~\cite{maaz2024video,lin2024video,cheng2024videollama,li2024llava}, encode sampled frames and project their features into the language-model input. The visual interface determines both what evidence the model receives and how that evidence is organized for reasoning. Dense interfaces pass thousands of visual tokens per video. LLaVA-OneVision, for example, uses 6,272 visual tokens for 32 frames, and the count grows with every added frame. Token selection and merging reduce how many visual tokens reach the language model~\cite{li2024llama,shen2024longvu,yang2025visionzip}; we instead focus on what each token represents.

\begin{figure*}[t]
  \centering
  \includegraphics[width=\linewidth]{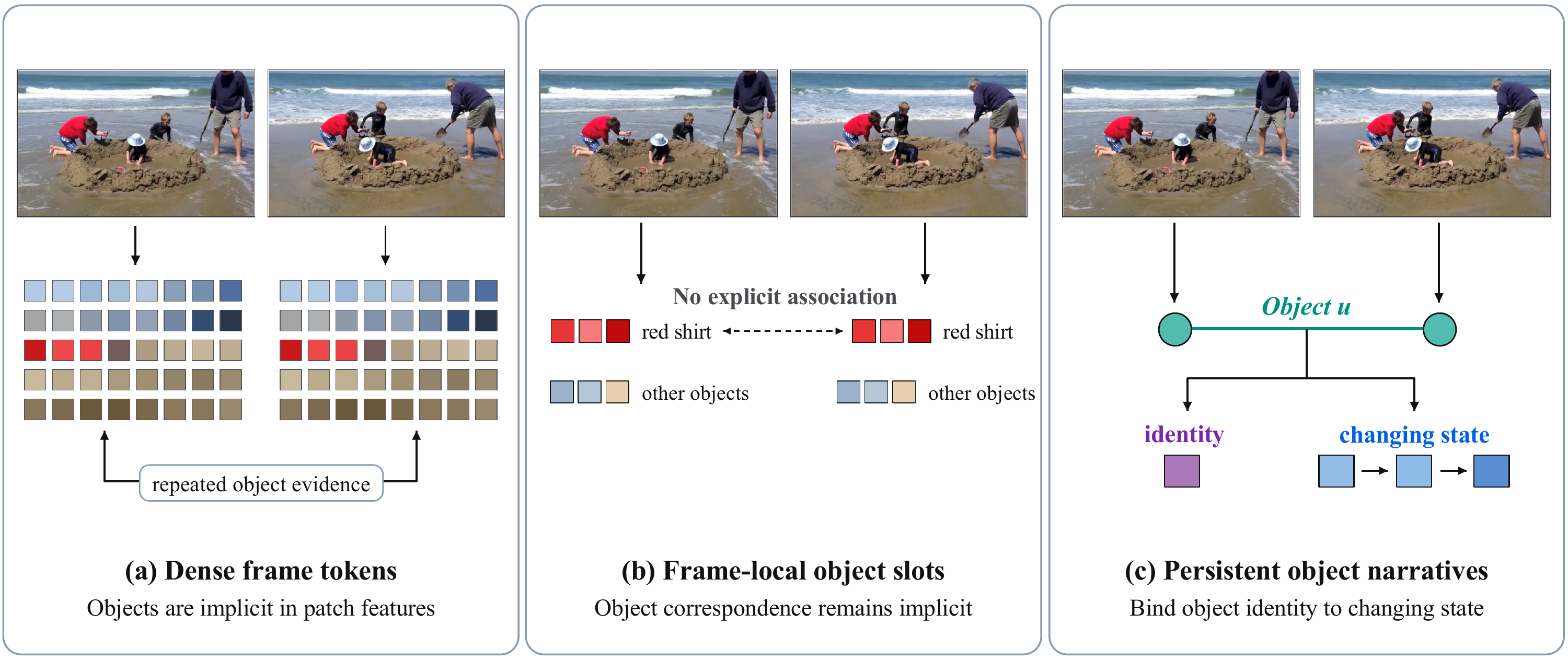}
  \caption{\textbf{Conceptual comparison of visual interfaces.} The same frames illustrate (a) dense frame tokens, (b) frame-local object slots, and (c) persistent object narratives. Panel (c) explicitly links the observations of object $u$ and represents its identity and changing state with separate token types.}
  \label{fig:overview}
\end{figure*}

Many video questions connect \emph{which object is involved}, \emph{what happens to it}, and \emph{when}. To answer what a person in a red shirt does later in a video, a model must know which observations in different frames show that person. Figure~\ref{fig:overview} illustrates this requirement. In (a), evidence about the person is spread over the patch tokens of each frame, with no explicit reference linking them. In (b), each frame is grouped into \emph{slots}: vectors that collect patches from object-like regions. The slots separate the person from other objects, but their correspondence across frames remains implicit (dashed link), so the language model must recover it from the tokens themselves. The problem is harder after an observation gap: an object that is occluded, briefly missed, or out of view may reappear in a different slot.

Object-centric learning provides spatial groups through Slot Attention~\cite{locatello2020object}, and video extensions add recurrence and temporal consistency~\cite{kipf2021conditional,zadaianchuk2023object,manasyan2025temporally}. Slot-VLM brings object and event slots into video-language modeling~\cite{xu2024slot}, and VideoOrion organizes object evidence with detection and tracking~\cite{feng2025videoorion}. We address two problems that these works leave open: linking learned slot observations into clip-level object histories without an external tracker, and presenting each history to the language model so that its appearance and its changing states stay connected. The two kinds of evidence are complementary: appearance identifies an object but not how it changes, and per-frame states describe changes but not which object they belong to.

We introduce \ours{}, which represents a video as \emph{persistent object narratives}. A narrative is the time-ordered set of observations assigned to one clip-level entry. The entry is kept while its object is out of view, so later observations can rejoin it. Panel (c) of Figure~\ref{fig:overview} shows this organization: the person's observations are assigned to object $u$, whose identity token (purple) summarizes its appearance and whose state tokens (blue) record how it changes. Recurrent slots produce the observations. A training-free memory links them: when a slot becomes visible again, memory compares it with stored entries by appearance, trajectory, and position, and either reattaches it to an existing entry or opens a new one. A typed tokenizer then turns each narrative into the two token types. Shared object embeddings tie an object's identity token to its state tokens, and segment embeddings mark when each state occurs. The visual sequence is thus indexed by objects, not frames.

The interface is compact and inexpensive to train. Its length is set by object capacity and segment count rather than by the number of frames. Its cached key/value context occupies 7.9 MiB per video, compared with 343 MiB for a 6,272-token dense input. Both backbones stay frozen. Language alignment trains 34.4M interface parameters, and no training stage uses object-level annotations.

We evaluate \ours{} on three open-ended VideoQA benchmarks. With 144 visual tokens per video, it achieves 75.6\%, 69.8\%, and 55.2\% accuracy on MSVD-QA, MSRVTT-QA, and ActivityNet-QA, higher reported accuracy than Slot-VLM on all three with 25\% fewer visual tokens. With the same backbones, input frames, and Stage III recipe, average pooling to 144 tokens is 6.9 points lower on ActivityNet-QA. Controlled ablations show that memory adds 2.8 to 4.2 points at the same token budget and that identity and state tokens provide complementary evidence: removing either type lowers ActivityNet-QA accuracy by 6.2 to 7.9 points.

Our contributions are summarized as follows:
\begin{itemize}
\item We introduce \ours{}, an object-centric Video-LLM that organizes learned slot observations into \emph{persistent object narratives} through training-free association.
\item We propose a \emph{typed object-state interface} that represents each narrative with one clip-level identity token and several segment-level state tokens, tied by shared object embeddings. The two token types are complementary.
\item We demonstrate a favorable accuracy--visual-token trade-off on three open-ended VideoQA benchmarks and analyze the contributions of persistent association, identity/state token content, and visual-token budget.
\end{itemize}

\begin{figure*}[t]
  \centering
  \includegraphics[width=\linewidth]{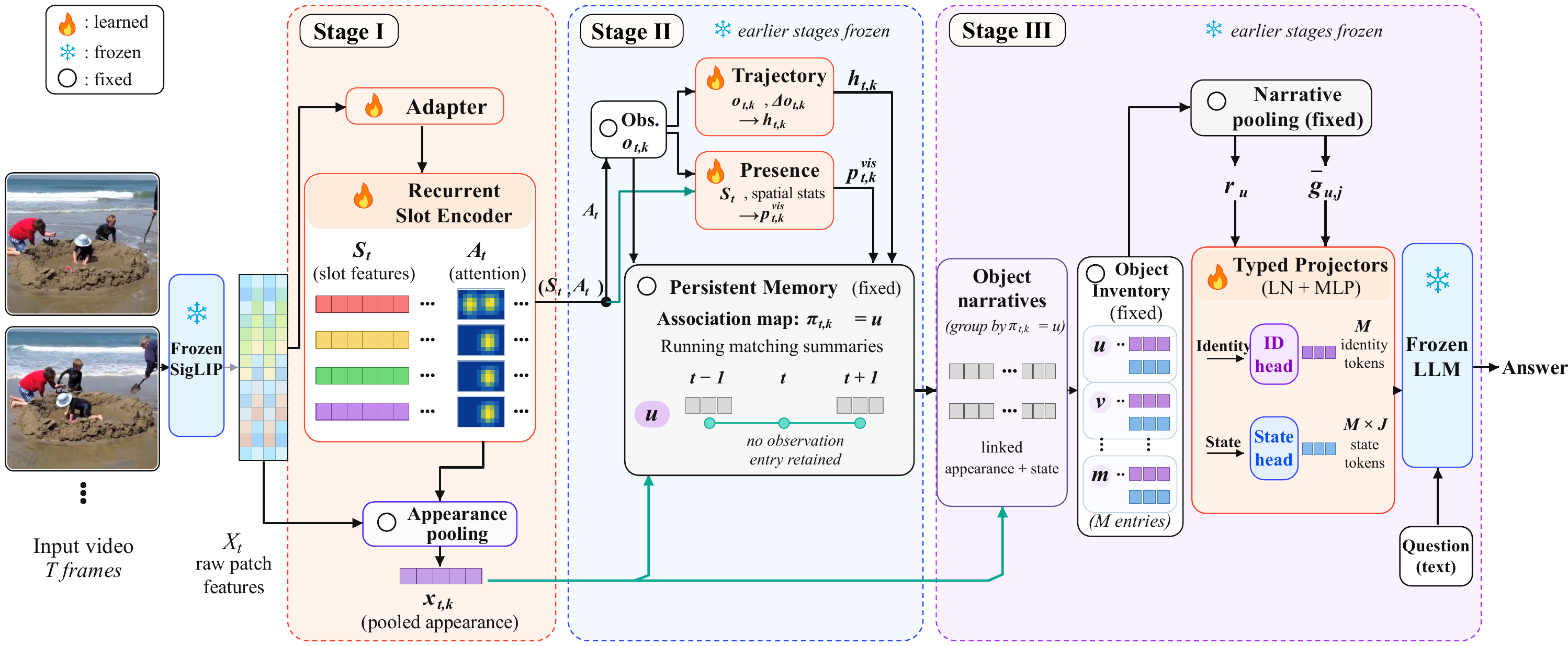}
  \caption{\textbf{The \ours{} pipeline.} Slots form frame-level observations, memory associates them into object narratives, and inventory selection and narrative pooling prepare the identity/state inputs. Dashed regions mark training stages. Flames denote learned modules, snowflakes frozen modules, and circles fixed operations. The memory inset illustrates entry retention across an observation gap. Obs. also uses consecutive SigLIP grids to estimate motion.}
  \label{fig:framework}
\end{figure*}

\section{Related Work}
\label{sec:related_work}

\paragraph{Video-LLMs and visual interfaces.}
Video-ChatGPT, Video-LLaVA, Video-LLaMA, and Chat-UniVi extend language models to video instruction following~\cite{maaz2024video,lin2024video,zhang2023video,jin2024chat}. VideoLLaMA~2, LLaVA-OneVision, and LLaVA-Video further improve video-language modeling through stronger architectures and training~\cite{cheng2024videollama,li2024llava,zhang2024llava}. Most interfaces construct visual sequences from sampled-frame features. Query bottlenecks~\cite{li2023blip,dai2023instructblip}, token selection and merging~\cite{bolya2022token,ryoo2021tokenlearner,chen2024fastv,yang2025visionzip}, and video-specific compression~\cite{li2024llama,shen2024longvu,xu2024slowfast,song2024moviechat,ma2024vista} reduce their input cost. Matryoshka Multimodal Models use nested token sets~\cite{cai2025matryoshka}, LLaVA-Mini pre-fuses visual information into text~\cite{zhang2025llavamini}, and recent methods coordinate temporal and spatial token reduction~\cite{shen2025fastvid,shao2025holitom,du2026unified}. Our focus is the semantic organization of the visual interface: representing recurring objects, their states, and the links between them.

\paragraph{Object-centric video learning.}
MONet, IODINE, and Slot Attention learn object-like decompositions~\cite{burgess2019monet,greff2019multi,locatello2020object}, while DINOSAUR applies slot grouping to pretrained visual features~\cite{seitzer2022bridging}. Video models propagate object representations~\cite{kipf2021conditional,zadaianchuk2023object}, learn temporal consistency~\cite{manasyan2025temporally}, and combine complementary representation objectives~\cite{seong2026synergistic}. SlotFormer models object-centric dynamics~\cite{wu2023slotformer}, SlotMatch distills object representations~\cite{grigore2025slotmatch}, and Dual-State Slot Attention separates per-frame appearance from persistent identity within the slot encoder~\cite{tran2026dualstate}. These methods provide the spatial groups and local continuity on which we build. We further link observations into clip-level entries and turn each entry into language-model tokens.

\paragraph{Object-aware video-language modeling.}
Slot-VLM combines object and event slots for video-language understanding~\cite{xu2024slot}. VideoOrion constructs object representations with a detection, segmentation, and tracking pipeline~\cite{feng2025videoorion}. One Trajectory, One Token uses panoptic sub-object trajectories~\cite{zheng2025trajectory}, and TrajTok constructs trajectory tokens from unified spatiotemporal segmentation~\cite{zheng2026trajtok}. These approaches show the value of organizing video information around entities. SORT associates detections using motion, and DeepSORT adds appearance features~\cite{bewley2016simple,wojke2017simple}. \ours{} applies this kind of association to learned slots rather than detections and represents each resulting history with a clip-level appearance token and segment-level state tokens tied by a shared object index.

\section{Method}
\label{sec:method}

\begin{figure*}[t]
  \centering
  \includegraphics[width=\linewidth]{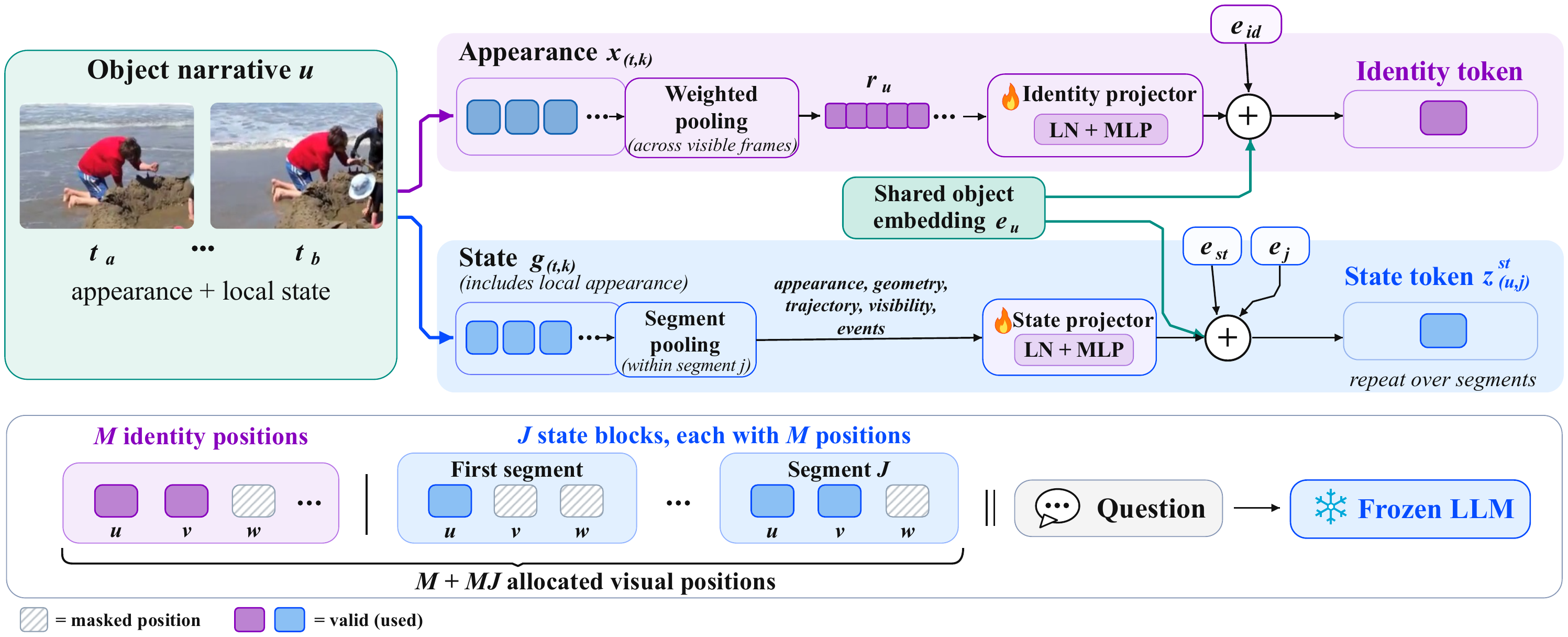}
  \caption{\textbf{Constructing identity and state tokens.} The upper purple path pools clip-level appearance, and the lower blue path pools segment-level state. The shared object embedding $e_u$ links the two token types. The bottom strip places $M$ identity positions before $J$ state blocks of $M$ positions each. Hatching marks an unused entry $w$ and an unobserved segment of entry $v$.}
  \label{fig:typed_interface}
\end{figure*}

Given a video $V$ and a question $q$, \ours{} builds a visual-token sequence $\mathcal Z(V)$ from the video alone and passes it, followed by the question, to a frozen language model.

\subsection{Framework Overview}
\label{sec:principle}

Figure~\ref{fig:framework} shows the pipeline. A frozen SigLIP encoder extracts frame features. A recurrent slot encoder groups them into object-like observations and pools each observation's appearance from the raw features. Lightweight modules add local descriptors, motion history, and observation confidence (Section~\ref{sec:observations}). A training-free memory links observations into clip-level entries; the time-ordered observations of an entry form its persistent object narrative (Section~\ref{sec:memory}). An object inventory keeps at most $M$ narratives, and typed projectors turn each into one identity token and several segment-level state tokens for the frozen LLM (Section~\ref{sec:tokens}).

\subsection{Object-Centric Visual Encoding}
\label{sec:observations}

For each of the $T$ sampled frames, a frozen SigLIP encoder~\cite{zhai2023sigmoid} extracts patch features $X_t\in\mathbb R^{P\times D_v}$. A feature adapter $\phi$ and a recurrent slot encoder $C$ group the $P$ patches into $K$ slots, each with a vector and an attention map over the patches:
\begin{equation}
 (S_t,A_t)=C(\phi(X_t);\widetilde S_{t-1}),\qquad
 \widetilde S_t=\Psi(S_t).
 \label{eq:slot_encoder}
\end{equation}
Here $S_t\in\mathbb R^{K\times D_s}$ holds the slot vectors and $A_t\in[0,1]^{K\times P}$ the attention maps. We call each frame-slot pair $(t,k)$ an \emph{observation} and $k$ its \emph{raw slot index}. A shot is a contiguous frame sequence bounded by scene cuts or video endpoints. Within a shot, the predictor $\Psi$ initializes the slots of the next frame; the first frame of a shot uses learned initial slots. Each attention map pools the appearance of its group from the raw SigLIP features:
\begin{equation}
 x_{t,k}=\sum_{p=1}^{P}A_{t,k,p}X_{t,p},\qquad
 \sum_p A_{t,k,p}=1.
 \label{eq:appearance}
\end{equation}
Adapted features decide how patches are grouped. Raw backbone features describe what each group shows.

Three lightweight modules describe each observation locally. The observation extractor derives spatial and shape statistics from the attention map and motion from consecutive SigLIP grids, giving $o_{t,k}\in\mathbb R^{25}$. A causal temporal encoder summarizes recent observations as a trajectory state $h_{t,k}\in\mathbb R^{64}$. A presence estimator predicts two confidence gates from slot vectors and spatial statistics. The existence gate $p^{\mathrm{exist}}$ controls whether an entry is created or cleared, and the visibility gate $p^{\mathrm{vis}}$ controls whether an observation updates its entry. A gate is on when its value exceeds $0.5$. Slot recurrence and motion history restart at each scene cut, detected from HSV-histogram changes. Appendix~\ref{app:representation_memory} details the descriptors, the encoder, and the gate targets.

\subsection{Persistent Object Memory}
\label{sec:memory}

Memory assigns observations to \emph{entries}. An entry is a clip-level record of associated observations: running summaries of its slot vector and appearance, a trajectory state, its recent observations, its position, and its last observation time. The assignment $\pi_{t,k}=u$ records that observation $(t,k)$ belongs to entry $u$. Memory has no trainable parameters and uses no tracking supervision.

\begin{figure}[!t]
  \centering
  \includegraphics[width=\columnwidth]{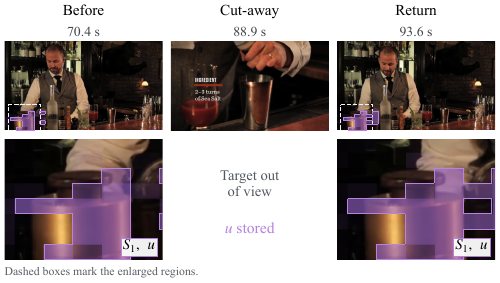}
  \caption{\textbf{Entry retention across a cut-away.} The lidded container on the left leaves the view while the camera cuts to a close-up. Its entry $u$ is kept without observations and receives the container again when the wide view returns. Purple shows model attention; dashed regions are enlarged below, labeled with their slot and entry. Appendix~\ref{app:direct_evaluation} shows full frames with pseudo-reference masks.}
  \label{fig:physical_reappearance}
\end{figure}

Entries persist through observation gaps. When an object is no longer observed, its entry is kept unchanged and can receive later observations, as in the memory inset of Figure~\ref{fig:framework} and the example of Figure~\ref{fig:physical_reappearance}. This retention is what makes the narratives persistent.

Recurrence carries each object while it stays visible; memory decides where a reappearing slot belongs. Within a shot, a continuously visible slot keeps its assignment while the existence gate is on. Matching runs when a slot becomes visible again: the slot is compared with each stored entry that no continuously visible slot holds. For the entry at memory location $m$, the score combines five cues:
\begin{equation}
 \rho_{t,k,m}=
 \frac{\sum_{\ell\in\mathcal C}w_\ell\gamma_\ell(t,k,m)}
 {\max(\sum_{\ell\in\mathcal C}w_\ell,\varepsilon)}.
 \label{eq:association}
\end{equation}
Here $\mathcal C$ contains slot-feature, pooled-appearance, trajectory-state, recent-observation, and position agreement; $w_\ell$ are fixed nonnegative weights and $\varepsilon=10^{-8}$. The feature cues use cosine similarity, and the position cue compares the current centroid with a velocity-based prediction. At a scene cut, all visible slots are re-matched by appearance alone.

Greedy matching accepts pairs scoring above a fixed threshold of 0.3 and uses each candidate location at most once. A matched slot takes over the entry's identifier and blends its trajectory state with the stored one. An unmatched observation that passes both gates starts a new entry. Visible observations update their entry's summaries. The observations assigned to one entry, ordered in time, form its object narrative. Appendix~\ref{app:representation_memory} gives the update rules and matching parameters; all QA results use the same values.

\subsection{Typed Object-State Tokens}
\label{sec:tokens}

Each narrative receives two token types: one identity token as a stable appearance reference, and state tokens that describe the object segment by segment. A shared object embedding marks which states belong to which identity.

\paragraph{Object inventory.}
The inventory keeps at most $M$ entries, ranked by how long they are observed and how much their state changes and weighted by objectness. Kept entries are renumbered as $u\in\mathcal U\subseteq\{1,\ldots,M\}$, and $\pi$ denotes the renumbered assignments. The inventory does not depend on the question, so a video's tokens are computed once and reused for all of its questions.

\paragraph{Identity summary.}
The identity path (upper path in Figure~\ref{fig:typed_interface}) pools appearance over all observations of entry $u$:
\begin{equation}
 r_u=\frac{\sum_{t,k:\pi_{t,k}=u}\alpha_{t,k}x_{t,k}}
 {\max(\sum_{t,k:\pi_{t,k}=u}\alpha_{t,k},\varepsilon)}.
 \label{eq:identity_pool}
\end{equation}
The weight $\alpha_{t,k}=p^{\mathrm{vis}}_{t,k}\max(\eta_{t,k},0.1)$ combines observation confidence with clip-normalized objectness $\eta_{t,k}\in[0,1]$. Appendix~\ref{app:objectness} defines objectness and inventory ranking.

\paragraph{Segment-state summaries.}
The state path (lower path) summarizes local evidence within each of $J$ uniform temporal segments. Each observation contributes a state feature
\begin{equation}
 g_{t,k}=[h^{\mathrm{mem}}_{t,k};o_{t,k};p^{\mathrm{vis}}_{t,k};b_{t,k};x_{t,k}],
 \label{eq:state}
\end{equation}
where $h^{\mathrm{mem}}_{t,k}$ is the trajectory state returned by memory and $b_{t,k}\in\{0,1\}^2$ flags an accepted match and a visibility-gate change. Keeping the local appearance $x_{t,k}$ lets state tokens describe visual changes as well as motion and geometry.

Observations of the same entry within a frame are averaged with normalized $\alpha$ weights, giving $\widehat g_{t,u}$. Let $\mathcal T_{u,j}$ be the frames of segment $j$ in which entry $u$ is observed. Its segment summary is
\begin{equation}
 \bar g_{u,j}=\frac{1}{|\mathcal T_{u,j}|}
 \sum_{t\in\mathcal T_{u,j}}\widehat g_{t,u},
 \label{eq:state_pool}
\end{equation}
computed only when $\mathcal T_{u,j}$ is nonempty; unobserved segments are masked.

\paragraph{Projection and shared indices.}
Separate layer normalizations (LN) and multilayer perceptrons (MLPs) $f_{\mathrm{id}}$ and $f_{\mathrm{st}}$ map the summaries into the language-model embedding space:
\begin{align}
 z^{\mathrm{id}}_u&=f_{\mathrm{id}}(\mathrm{LN}_{\mathrm{id}}(r_u))+e_{\mathrm{id}}+e_u,
 \label{eq:identity_token}\\
 z^{\mathrm{st}}_{u,j}&=f_{\mathrm{st}}(\mathrm{LN}_{\mathrm{st}}(\bar g_{u,j}))+e_{\mathrm{st}}+e_u+e_j.
 \label{eq:state_token}
\end{align}
The type embeddings $e_{\mathrm{id}}$ and $e_{\mathrm{st}}$ mark the two roles, and $e_j$ marks the temporal segment. The shared object embedding $e_u$ is added to both token types of entry $u$, tying its appearance reference to its states. Object indices come from the memory entries of each video, so observations assigned to one entry share an index across raw slots.

\paragraph{Input sequence.}
The sequence places the $M$ identity positions before $J$ blocks of $M$ state positions (bottom of Figure~\ref{fig:typed_interface}). Under causal attention, every valid state token can thus attend to all valid identity tokens. Unused entries and unobserved segments are masked. All $M(1+J)$ positions, masked or not, precede the question and count as visual tokens. Appendix~\ref{app:typed_interface} gives the masking and empty-inventory rules.

\begin{table*}[t]
  \centering
\caption{\textbf{Open-ended VideoQA results.} All methods use a 7B-scale language model. Acc. is accuracy (\%), Score is answer quality (0--5), and \tok{} is the number of visual tokens (GOP: group of pictures). Instruction data is the approximate number of samples in the instruction-tuning stage (V: video only; Mix: mixed modalities). Baseline results and data sizes are from the original papers. $\dagger$: benchmark-related training data; $\ddagger$: derived token count; --: not reported.}
  \label{tab:main_results}
  \setlength{\tabcolsep}{4pt}
  {\small
  \begin{tabular}{@{}lcrcccccc@{}}
    \toprule
    & & & \multicolumn{2}{c}{MSVD-QA} & \multicolumn{2}{c}{MSRVTT-QA} &
    \multicolumn{2}{c}{ActivityNet-QA} \\
    \cmidrule(lr){4-5}\cmidrule(lr){6-7}\cmidrule(l){8-9}
    Method & \shortstack{Instruction\\data} & \tok{} & Acc. & Score & Acc. & Score & Acc. & Score \\
    \midrule
    ReMoRa$^{\ddagger}$ & 200K V & 729/GOP & 73.1 & 4.0 & -- & -- & 60.5 & 3.7 \\
    LLaMA-VID & 763K Mix & 2/frame & 69.7 & 3.7 & 57.7 & 3.2 & 47.4 & 3.3 \\
    Video-LLaVA & 765K Mix & 2,048 & 70.7 & 3.9 & 59.2 & 3.5 & 45.3 & 3.3 \\
    LLaVA-OneVision & 1.6M Mix & 6,272 & -- & -- & -- & -- & 56.6 & -- \\
    LLaVA-ST$^{\ddagger}$ & 2.36M Mix & 2,520 & 75.9 & 4.1 & 59.0 & 3.5 & -- & -- \\
    EMA & 2.4M Mix & 648 & 75.8 & 4.1 & 58.5 & 3.5 & 52.1 & 3.5 \\
    LLaVA-Video$^{\dagger\ddagger}$ & 2.7M Mix & 10,816 & -- & -- & -- & -- & 56.5 & -- \\
    VideoOrion+ & $\sim$4M Mix & -- & -- & -- & -- & -- & 60.3 & 3.7 \\
    \midrule
    Video-ChatGPT & 100K V & 356 & 64.9 & 3.3 & 49.3 & 2.8 & 35.2 & 2.8 \\
    Slot-VLM & 100K V & 192 & 74.9 & 3.8 & 69.6 & 3.4 & 48.3 & 3.4 \\
    \textbf{\ours{}} & \textbf{100K V} & \textbf{144} & \textbf{75.6} & \textbf{3.8} &
    \textbf{69.8} & \textbf{3.4} & \textbf{55.2} & \textbf{3.6} \\
    \bottomrule
  \end{tabular}}
\end{table*}

\subsection{Training Strategy}
\label{sec:training}

Training follows the three stages in Figure~\ref{fig:framework}. \textbf{Stage I} learns the adapter, recurrent slots, predictor, and reconstruction decoder with feature reconstruction, SlotContrast (a contrastive loss for temporally consistent slots), and Synergistic Representation Learning (SRL)~\cite{seitzer2022bridging,manasyan2025temporally,seong2026synergistic}. \textbf{Stage II} freezes these modules and learns trajectory and presence estimation from future-observation prediction and attention-derived confidence targets. \textbf{Stage III} freezes the visual modules and learns the typed projectors, layer normalizations, and type/object/segment embeddings from video instruction conversations. SigLIP and Qwen2-7B-Instruct~\cite{yang2024qwen2technical} stay frozen throughout; memory, inventory selection, and pooling have no trainable parameters.

For an answer $a_1,\ldots,a_L$, Stage III minimizes answer-token cross-entropy:
\begin{equation}
\begin{split}
 \mathcal L_{\mathrm{QA}}(\theta)
 =-\frac{1}{L}\sum_{\ell=1}^{L}
 \log p_{\mathrm{LLM}}\bigl(
 a_\ell\mid\mathcal Z_\theta(V),q,a_{<\ell}\bigr),
\end{split}
\label{eq:qa_alignment}
\end{equation}
where $\theta$ denotes the interface parameters; gradients flow through the frozen language model into them. Appendix~\ref{app:training} lists all training settings.

\section{Experiments}
\label{sec:experiments}

We first compare \ours{} with existing Video-LLMs and compact visual interfaces. We then analyze persistent narratives and the typed interface through controlled ablations, which share one training and evaluation protocol. Qualitative examples and an efficiency analysis follow.

\subsection{Experimental Setup}
\label{sec:setup}

\paragraph{Training data and benchmarks.}
Stages I--II use the ActivityNet videos of VideoInstruct-100K~\cite{maaz2024video} without text, and Stage III uses its instruction conversations. We evaluate on the official test sets of MSVD-QA, MSRVTT-QA~\cite{xu2017video}, and ActivityNet-QA~\cite{yu2019activitynet}. MSVD-QA and MSRVTT-QA contain short web videos, and ActivityNet-QA contains longer untrimmed activities. The training videos do not overlap with the ActivityNet-QA test videos.

\paragraph{Evaluation metrics.}
Following Slot-VLM~\cite{xu2024slot}, we use the Video-ChatGPT evaluation protocol~\cite{maaz2024video}. The judge, gpt-3.5-turbo, compares each generated answer with the reference and returns a yes/no correctness decision and a 0--5 quality score. Acc. is the percentage judged correct, and Score is the mean quality score.

\paragraph{Implementation details.}
We use frozen SigLIP and Qwen2-7B-Instruct~\cite{zhai2023sigmoid,yang2024qwen2technical} as the visual and language backbones. Each video is uniformly sampled into 100 frames at $384\times384$ resolution. The slot encoder uses $K=11$ slots, the inventory keeps at most $M=16$ entries, and each narrative is divided into $J=8$ segments, giving $M(1+J)=144$ visual tokens. Answers are decoded greedily with at most 64 new tokens.

\subsection{VideoQA Results}
\label{sec:main_results}

Table~\ref{tab:main_results} compares \ours{} with general Video-LLMs~\cite{li2024llama,maaz2024video,lin2024video,li2024llava,zhang2024llava,li2025llavast}, object-based interfaces~\cite{xu2024slot,feng2025videoorion}, and motion-aware methods~\cite{zhao2025ema,yashima2026remora}. The lower block uses 100K video instruction samples; the upper block uses about 2 to 40 times as many, often mixing image and video data. Within the 100K-data block, \ours{} has the highest reported accuracy on all three benchmarks and uses 25\% fewer visual tokens than Slot-VLM, the closest slot-based model. Appendix~\ref{app:additional_results} reports ActivityNet-QA accuracy by question type.

\begin{table}[t]
  \centering
\caption{\textbf{Compact visual interfaces on ActivityNet-QA.} The upper block builds on LLaVA-OneVision-7B; its dense result and the four 2\%-retention results are reported by Du et al.~\cite{du2026unified} under one setup (the original LLaVA-OneVision report gives 56.6\%). The lower block shares our backbones, 100 input frames, and Stage III training; $\ddagger$ average-pools the SigLIP features on a $16\times3\times3$ spatiotemporal grid (Appendix~\ref{app:reproducibility_settings}).}
  \label{tab:compact_comparison}
  \setlength{\tabcolsep}{4pt}
  {\small
  \begin{tabular}{@{}lrrr@{}}
    \toprule
    Interface & \tok{} & Acc. & Score \\
    \midrule
    LLaVA-OneVision & 6,272 & 54.5 & 3.6 \\
    + VisionZip (2\%) & $\sim$125 & 41.7 & 3.1 \\
    + FastVID (2\%) & $\sim$125 & 49.6 & 3.4 \\
    + HoliTom (2\%) & $\sim$125 & 49.9 & 3.3 \\
    + Unified (2\%) & $\sim$125 & 50.2 & 3.4 \\
    \midrule
    Avg. pooling$^{\ddagger}$ & 144 & 48.3 & 3.1 \\
    \textbf{\ours{}} & \textbf{144} & \textbf{55.2} & \textbf{3.6} \\
    \bottomrule
  \end{tabular}}
\end{table}

\begin{figure*}[!t]
  \centering
  \includegraphics[width=\linewidth]{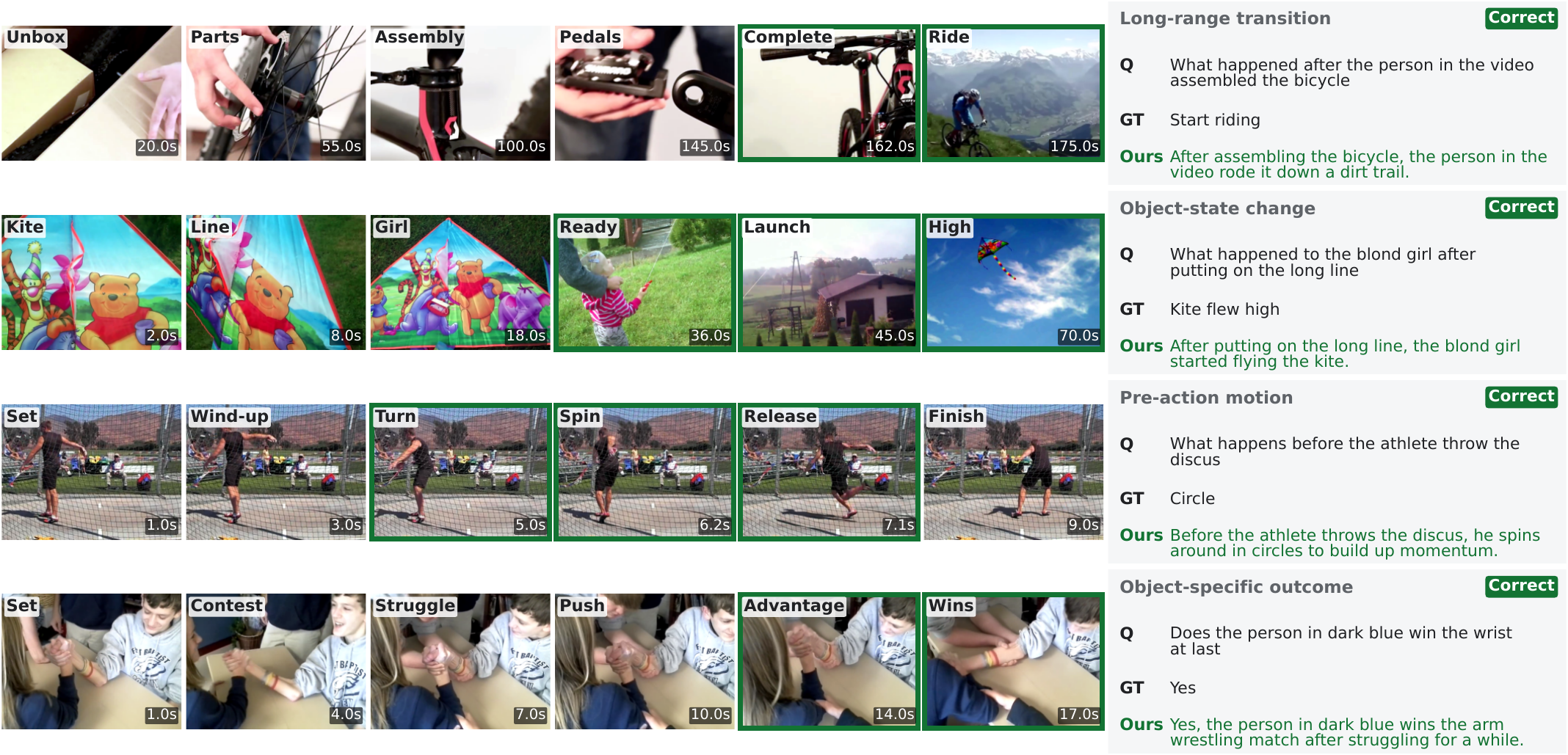}
  \caption{\textbf{Temporal and object-specific VideoQA examples.} Each row shows source frames, a question, the reference answer (GT), and our prediction. From top to bottom: a long-range transition, an object-state change, pre-action motion, and an outcome for a specified person. Green borders mark displayed answer-relevant frames.}
  \label{fig:qualitative_main}
\end{figure*}

\paragraph{Compact visual interfaces.}
Table~\ref{tab:compact_comparison} compares interfaces of about 144 tokens on ActivityNet-QA. \ours{} exceeds all compressed LLaVA-OneVision variants and is close to its dense input. With our backbones, frames, training data, and token budget, average pooling reaches 48.3\%, 6.9 points lower, so the gap comes from the interface.

\subsection{Persistent Object Narratives}
\label{sec:narrative_analysis}

Table~\ref{tab:model_variants} builds persistent narratives step by step. All rows share the identity/state tokenizer, the $M=16,J=8$ budget, and Stage III training. \emph{Slot observations} use raw slot indices as entries and omit the trajectory state. Starting from them, \emph{Trajectory only} adds the trajectory state to state tokens, and \emph{Memory only} instead adds memory; without a trajectory state, its matching omits that cue. The full model uses both.

\begin{table}[t]
  \centering
\caption{\textbf{Building persistent narratives.} VideoQA accuracy (\%) with 144 visual tokens. The first two rows use raw slot indices as entries; the last two use memory entries.}
  \label{tab:model_variants}
  {\small
  \setlength{\tabcolsep}{3.2pt}
  \begin{tabular}{@{}lrrr@{}}
    \toprule
    Configuration & MSVD & MSRVTT & ActivityNet \\
    \midrule
    Slot observations & 71.0 & 65.0 & 50.3 \\
    Trajectory only & 72.8 & 65.6 & 52.4 \\
    Memory only & 74.2 & 67.1 & 52.8 \\
    Full \ours{} & 75.6 & 69.8 & 55.2 \\
    \bottomrule
  \end{tabular}}
\end{table}

Each component improves accuracy on all three benchmarks, and the full model gains 4.6, 4.8, and 4.9 points over slot observations on MSVD-QA, MSRVTT-QA, and ActivityNet-QA. On ActivityNet-QA, slot observations already exceed same-recipe average pooling (50.3\% vs.\ 48.3\%; Table~\ref{tab:compact_comparison}). The trajectory state also improves linear prediction of future observations (Appendix~\ref{app:additional_results}). Comparing Trajectory only with the full model measures the effect of memory: at the same token budget, memory adds 2.8, 4.2, and 2.8 points. Memory also consolidates observations. It attaches reappearing slots to existing entries and yields fewer entries and active positions on average than raw slot indices (6.3 to 9.0 vs.\ 8.3 to 9.5 entries per video; Appendix~\ref{app:representation_memory}).

\paragraph{Slot grouping.}
Learned slots group objects across frames more consistently than fixed spatial groups and temporally matched $k$-means clustering of SigLIP features (Appendix~\ref{app:direct_evaluation}). In video foreground adjusted Rand index (FG-ARI), they reach .222 versus .149 and .081 on 50 ActivityNet videos with SAM~3~\cite{carion2025sam3} pseudo-reference masks, and .135 versus .061 and .022 on 90 DAVIS-2017 videos with human masks~\cite{ponttuset2017davis}. Narratives depend on this consistency: removing SlotContrast from Stage I lowers ActivityNet-QA accuracy from 55.2\% to 50.7\% (Appendix~\ref{app:additional_results}).

\paragraph{Links across time.}
A \emph{link} joins two observations from different frames that share an entry~\cite{amigo2009comparison}. We evaluate links on 40 ActivityNet videos, where SAM~3 tracks objects named by Qwen3-VL~\cite{qwen3vl8b}. Memory raises identity F1 (IDF1) from 64.4\% with raw slot indices to 67.0\%. It keeps 95.0\% of same-object links across observation gaps within a raw slot, compared with 32.0\% for matching on pooled appearance alone, and it links objects that return in a different raw slot, which raw indices cannot. Figure~\ref{fig:physical_reappearance} shows an entry kept through a cut-away. The learned slot features carry most of this association: removing slot-feature similarity from matching lowers IDF1 by 5.6 points, and removing pooled appearance removes about two thirds of the links across raw slots (Appendix~\ref{app:direct_evaluation}).

\subsection{Typed Object-State Interface}
\label{sec:interface_analysis}

Table~\ref{tab:typed_ablation} tests the two paths of Figure~\ref{fig:typed_interface} on ActivityNet-QA. Identity tokens summarize an object over the whole clip, and state tokens describe it segment by segment. The two are complementary: removing either type lowers accuracy by 6 to 8 points. The loss remains when the 144-token budget is kept by repeating identity tokens or by adding a ninth state segment. Identity tokens matter more than their number suggests: each is the clip-level reference that all state tokens of its object can attend to. The shared object embedding $e_u$ binds the two types; removing it costs 1.8 points. Building identity summaries from slot vectors instead of pooled SigLIP features costs 1.6 points, supporting the split in Section~\ref{sec:observations}: slots decide grouping, and backbone features describe content. Removing appearance from state tokens costs 1.2 points.

\begin{table}[t]
  \centering
\caption{\textbf{Identity/state interface ablations on ActivityNet-QA.} $\Delta$ is the accuracy change in percentage points from the full model. Indented rows keep 144 tokens with a ninth state segment or with identity tokens repeated in the state positions.}
  \label{tab:typed_ablation}
  \setlength{\tabcolsep}{4pt}
  {\small
  \begin{tabular}{@{}lrrr@{}}
    \toprule
    Variant & \tok{} & Acc. & $\Delta$ \\
    \midrule
    \textbf{Full model} & \textbf{144} & \textbf{55.2} & -- \\
    w/o identity tokens & 128 & 47.3 & -7.9 \\
    \quad state only, $J{=}9$ & 144 & 47.5 & -7.7 \\
    w/o state tokens & 16 & 49.0 & -6.2 \\
    \quad identity repeated & 144 & 49.3 & -5.9 \\
    w/o object embedding $e_u$ & 144 & 53.4 & -1.8 \\
    w/o segment embedding $e_j$ & 144 & 54.2 & -1.0 \\
    w/ slot-vector identity & 144 & 53.6 & -1.6 \\
    w/o state appearance & 144 & 54.0 & -1.2 \\
    \bottomrule
  \end{tabular}}
\end{table}

\begin{table}[t]
  \centering
\caption{\textbf{Object capacity and temporal granularity.} ActivityNet-QA results with $M$ object positions and $J$ temporal segments, allocating $M(1+J)$ visual tokens. Bold marks the default configuration.}
  \label{tab:budget_main}
  \setlength{\tabcolsep}{7pt}
  \begin{tabular}{@{}rrrr@{}}
    \toprule
    $M$ & $J$ & \tok{} & Acc. \\
    \midrule
    8 & 8 & 72 & 51.2 \\
    16 & 4 & 80 & 53.9 \\
    \textbf{16} & \textbf{8} & \textbf{144} & \textbf{55.2} \\
    24 & 8 & 216 & 55.8 \\
    16 & 16 & 272 & 55.5 \\
    \bottomrule
  \end{tabular}
\end{table}

\paragraph{Object capacity and temporal granularity.}
Table~\ref{tab:budget_main} varies object capacity $M$ and segment count $J$. At the default operating point, halving $M$ from 16 to 8 lowers accuracy by 4.0 points, whereas halving $J$ from 8 to 4 lowers it by 1.3. Reducing $M$ can drop whole narratives from the inventory, whereas reducing $J$ only coarsens the timeline of each narrative. Larger budgets add at most 0.6 points, so $M=16,J=8$ is a compact operating point.

\subsection{Qualitative Examples}
\label{sec:qualitative}

Figure~\ref{fig:qualitative_main} shows four correct answers to questions about object states. The first two connect an earlier state to a later event: the assembled bicycle is ridden, and the prepared kite flies. The third concerns motion before a discus throw, and the fourth the outcome for a specified person. Each question relates states of one object across time. Appendix~\ref{app:additional_results} shows failure cases.

\subsection{Efficiency Analysis}
\label{sec:efficiency}

We measure the language-model cost of visual context with frozen Qwen2-7B on an RTX~3090, feeding random visual embeddings of increasing length. At 144 positions, prefill takes 58.8 ms and the visual key/value (KV) cache occupies 7.9 MiB, compared with 1,414.9 ms and 343.0 MiB at 6,272 positions. The over 43-fold smaller cache matters when the visual context of many videos is kept for repeated questions. Going from 32 to 100 frames raises the measured LLaVA-OneVision prefill from 1.4 s to 5.2 s, whereas \ours{} stays at 144 positions and 0.06 s. The saving lies in the language model; visual encoding takes 5.2 s per 100-frame video, of which SigLIP takes 4.6 s and the object-centric modules, including memory, about 0.6 s, and it runs once per video. Appendix~\ref{app:runtime} reports end-to-end timings.
\section{Conclusion}
\label{sec:conclusion}

We presented \ours{}, a Video-LLM built on persistent object narratives. Learned slot grouping, training-free association, and a typed identity/state interface organize visual evidence around objects and their changes over time, and shared object embeddings connect each object's clip-level identity to its segment-level states. With 144 visual tokens, \ours{} is competitive on three open-ended VideoQA benchmarks, and controlled ablations show that persistent association and both token types contribute to accuracy. At a small token budget, organizing visual evidence into persistent object entries is an effective alternative to compressing frame tokens. Future work will improve observation selection and cross-slot association, explore finer temporal reasoning within the same object-state interface, and study the interface with larger-scale training.

\FloatBarrier

{\small
\bibliographystyle{unsrtnat}
\bibliography{references}
}

\appendix
\section{Object Representation and Memory}
\label{app:representation_memory}

The visual pipeline builds object observations and links them into clip-level entries. Slot attention provides spatial groups, trajectory and presence modules describe their local evolution, and persistent memory assigns them to entries. The resulting assignment map decides which observations form each narrative; Appendix~\ref{app:typed_interface} pools them into tokens. Observation extraction, appearance pooling, and the memory update rule have no trainable parameters.

\subsection{Feature Extraction and Slot Attention}
\label{app:feature_attention}

For a sequence of $T$ sampled frames, the frozen SigLIP encoder produces raw patch features $X_t\in\mathbb R^{P\times D_v}$ at each frame $t$, with $D_v=1152$ and $P=729$ at the $384\times384$ input resolution. A two-layer feature adapter maps these features to the input space of the recurrent slot encoder, which produces $K=11$ slots of dimension $D_s=256$. Slot attention on the adapted features decides \emph{which} patches belong together, and the raw SigLIP features describe \emph{what} those patches show.

We retain two normalizations of the slot-attention weights. For frame $t$, slot $k\in\{1,\ldots,K\}$, and patch $p\in\{1,\ldots,P\}$, we write $A_t,B_t\in[0,1]^{K\times P}$ for the patch-normalized and slot-competitive forms, respectively:
\begin{equation}
  \sum_{p=1}^{P}A_{t,k,p}=1,
  \qquad
  \sum_{k=1}^{K}B_{t,k,p}=1.
\end{equation}
$A_t$ is used for centroids, moments, appearance pooling, and object-state descriptors. $B_t$ estimates the fraction of the image assigned to a slot, $\mu_{t,k}=P^{-1}\sum_p B_{t,k,p}$. At a detected scene cut, the recurrent slots are reset to the learned initial slots.

\begin{figure*}[t]
  \centering
  \includegraphics[width=\textwidth]{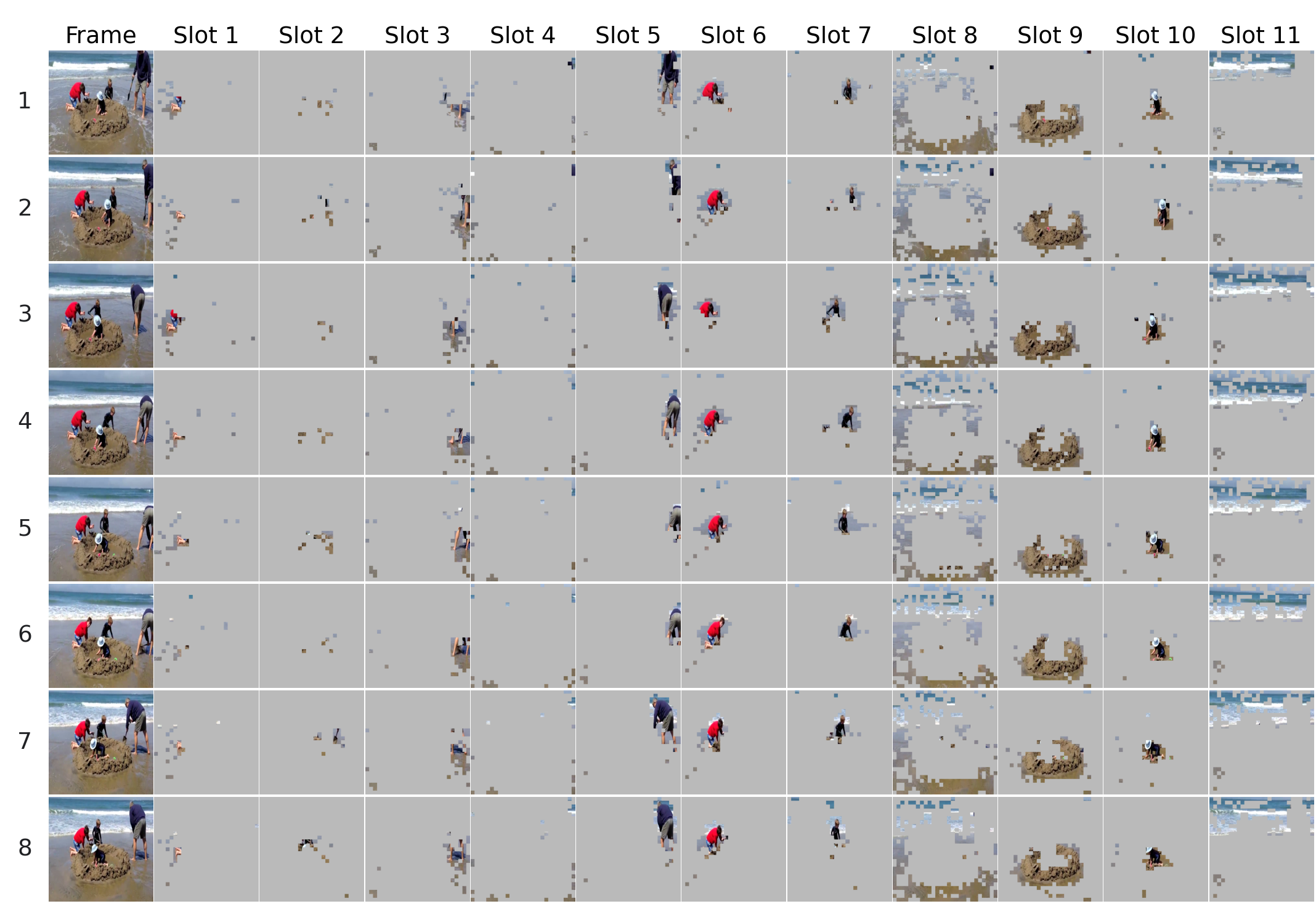}
  \caption{Spatial slot-attention masks over eight uniformly sampled frames from one shot. The first column shows the input frames; the remaining 11 columns show the patches assigned to each slot by the competitive attention $B_t$. Different slots consistently capture individual people, the sandcastle, and complementary shoreline regions.}
  \label{fig:supp_slot_attention_masks}
\end{figure*}

Reading Figure~\ref{fig:supp_slot_attention_masks} down a column follows one slot across the eight frames. Slot 6 groups the red-shirt person, and Slot 9 the sandcastle. These recurrent groups provide the observations passed to memory.

\subsection{Observation, Trajectory, and Presence}
\label{app:trajectory_state}

The parameter-free observation extractor forms $o_{t,k}\in\mathbb R^{D_o}$ with $D_o=25$ from three groups of statistics:
\begin{itemize}
\item \textbf{Spatial statistics (12 dimensions):} centroid, slot-competitive area $\mu_{t,k}$, three second moments, peak response, normalized entropy, and the attention mass touching the four image borders.
\item \textbf{Motion statistics (5 dimensions):} horizontal and vertical displacement, displacement variance, displacement magnitude, and feature change energy. Patch displacement is estimated by nearest-neighbor cosine matching between consecutive raw SigLIP feature grids and then pooled with the slot mask.
\item \textbf{Shape statistics (8 dimensions):} horizontal and vertical skewness and kurtosis, aspect ratio, horizontal and vertical profile peaks, and the ratio between the horizontal and vertical profile entropies. The profile peaks are log-scaled relative to a uniform profile.
\end{itemize}

The trajectory encoder receives the current observation together with its one- and two-frame lag differences, $[o_{t,k};o_{t,k}-o_{t-1,k};o_{t,k}-o_{t-2,k}]\in\mathbb R^{75}$, and maps them to $h_{t,k}\in\mathbb R^{D_h}$ with $D_h=64$ using a three-layer causal temporal convolution. Missing lag differences are set to zero. Each shot is encoded independently, so the differences and causal history restart at a scene boundary.

The presence estimator takes the slot vector and spatial statistics and predicts the two confidence gates $p^{\mathrm{exist}}$ and $p^{\mathrm{vis}}$, both supervised by attention concentration. Let $m_{t,k}=\max_p A_{t,k,p}$ be peak attention and $H_{t,k}=-(\log P)^{-1}\sum_p A_{t,k,p}\log A_{t,k,p}$ be normalized entropy, with the standard convention $0\log0=0$. We compute
\begin{equation}
  c_{t,k}=(1-H_{t,k})(Pm_{t,k}),
\end{equation}
and use $\mathbf{1}[c_{t,k}>3]$ and $\mathbf{1}[c_{t,k}>5]$ as the binary training targets for $p^{\mathrm{exist}}$ and $p^{\mathrm{vis}}$, respectively, where $\mathbf{1}[\cdot]$ is the indicator function. The gates thus estimate how concentrated a slot's attention is, which serves as observation confidence. The two targets threshold the same confidence at two levels: a slot keeps its identifier above the lower level and updates its entry only above the higher one. Appendix~\ref{app:association_design} varies the two thresholds.

\subsection{Training-Free Persistent Memory}
\label{app:persistent_memory}

The memory bank has $K$ locations, one per raw slot index, and each location stores at most one entry with its identifier. An entry stays at its location while its slot is invisible, so up to $K$ stored entries are candidates for re-association at any time. When a slot becomes visible and receives a new or transferred entry, that entry replaces the one stored at the slot's location, and the replaced identifier can no longer be matched. Replaced identifiers remain in the recorded assignments, so a clip can contain more than $K$ identifiers; Appendix~\ref{app:typed_interface} keeps at most $M$ of them.

Each entry stores exponential moving averages of the slot feature and the attention-pooled SigLIP appearance, the latest trajectory state, the mean of a five-observation buffer, the latest centroid and velocity, the last visible frame, and the entry identifier. The buffer is updated while the visibility gate is on and starts by repeating the first observation. Slot and appearance summaries support matching. The retained trajectory state also enters state tokens, which pool associated observation features (Appendix~\ref{app:typed_interface}).

Matching is triggered when a slot's visibility gate turns from off to on, and for every visible slot at a scene cut. Within a shot, a slot whose gate stays on keeps its assignment, since recurrent initialization carries it forward. At a cut, re-initialization breaks this continuity, so each visible slot releases its identifier and is matched again, with its previous entry among the candidates. Each slot to be matched is compared with each inactive entry using five signals: slot-feature, pooled-appearance, trajectory-state, and recent-observation cosine similarity, and Gaussian proximity to the velocity-predicted centroid. We index a candidate entry by the memory location $m$ that stores it. Let $\mathcal{C}$ index these signals, $\gamma_\ell(t,k,m)$ denote the value of signal $\ell$ between slot $(t,k)$ and location $m$, and $w_\ell\geq0$ its weight. Cosine cues lie in $[-1,1]$, and the Gaussian position cue lies in $[0,1]$. The matching score is
\begin{equation}
  \rho_{t,k,m}=
  \frac{\sum_{\ell\in\mathcal{C}}w_\ell\gamma_\ell(t,k,m)}
       {\max\!\left(\sum_{\ell\in\mathcal{C}}w_\ell,\varepsilon\right)},
  \qquad \varepsilon=10^{-8}.
\end{equation}
On ordinary frames, the weights for slot feature, pooled appearance, trajectory state, recent observation, and position are $(0.3,0.2,0.2,0.1,0.2)$, normalized by their sum in the score. The position signal uses a Gaussian bandwidth of $0.1$. When matching is triggered at a scene cut, motion cues are dropped and the score weights slot-feature and pooled-appearance similarity equally. The acceptance threshold is $0.3$.

Greedy one-to-one matching uses each candidate location at most once per step. An accepted match transfers the identifier and blends the stored and current trajectory states; if the entry came from another location, that location is cleared. An unmatched observation passing both gates starts a new identifier. Visible observations update the slot and appearance summaries with momentum $0.9$, and inactive entries are held. The map $\pi_{t,k}\in\{-1\}\cup\mathbb N_0$ records the assignments: nonnegative integers label entries, and $-1$ denotes no assignment.

\begin{table}[t]
\centering
\caption{Memory and inventory statistics over 520 MSVD, 2,990 MSRVTT, and 800 ActivityNet test videos. A trigger is an off-to-on transition of the visibility gate in a 100-frame clip. Entries are counted before inventory selection with $M=16$; active positions are the unmasked positions among the 144 allocated. The fallback never occurs under either assignment.}
\label{tab:supp_memory_triggers}
\setlength{\tabcolsep}{2.5pt}
{\small
\begin{tabular}{@{}>{\raggedright\arraybackslash}p{0.44\columnwidth}rrr@{}}
\toprule
 & MSVD & MSRVTT & ActivityNet \\
\midrule
Triggers per video & 32.6 & 35.7 & 74.9 \\
With a candidate (\%) & 95.5 & 96.0 & 97.6 \\
Accepted given a candidate (\%) & 96.8 & 95.5 & 95.1 \\
\midrule
\multicolumn{4}{@{}l}{\emph{Memory entries (full model)}} \\
Entries, mean / max & 6.3 / 11 & 6.7 / 15 & 9.0 / 20 \\
Videos with ${>}M$ entries (\%) & 0.0 & 0.0 & 1.5 \\
Active positions, mean & 46.5 & 46.8 & 55.3 \\
Active positions, p10--p90 & 35--58 & 36--58 & 43--67 \\
\midrule
\multicolumn{4}{@{}l}{\emph{Raw slot indices (Trajectory only)}} \\
Entries, mean / max & 8.3 / 10 & 9.0 / 10 & 9.5 / 10 \\
Active positions, mean & 52.1 & 54.6 & 67.5 \\
Active positions, p10--p90 & 39--65 & 42--67 & 52--81 \\
\midrule
Empty-inventory fallback (\%) & 0.0 & 0.0 & 0.0 \\
\bottomrule
\end{tabular}}
\end{table}

Table~\ref{tab:supp_memory_triggers} shows how often reappearance triggers matching on the three test sets. Most triggers follow a gap of a few sampled frames and return the slot to its own entry, so nearly all of them find and accept a candidate. More than $M$ entries occur in at most 1.5\% of videos. With raw slot indices as entries, as in Trajectory only, each raw slot that appears keeps one entry for the whole clip. One of the eleven slots collects the diffuse background, about a fifth of the patches, and never passes the visibility gate, so at most ten such entries arise. Memory attaches later slots to existing entries and forms fewer entries and active positions on average.

\paragraph{Matching cues and state updates.}
For numerical stability, feature similarity uses $\operatorname{cos}_\epsilon(a,b)=a^\top b/[\max(\|a\|_2,\epsilon)\max(\|b\|_2,\epsilon)]$ with $\epsilon=10^{-12}$. The four cosine comparisons pair the current slot, appearance, local trajectory, and observation with the corresponding stored summaries. For location $m$, let $\mathbf c_m$ be the stored centroid, $\mathbf v_m$ the velocity, and $\tau_m$ the last visible frame; let $\mathbf c_{t,k}$ be the current centroid. The position cue is
\begin{align}
 \widehat{\mathbf c}_{t,m}&=\operatorname{clip}_{[0,1]^2}
 \bigl(\mathbf c_m+\max(t-\tau_m,1)\mathbf v_m\bigr),\\
 \gamma_{\mathrm{pos}}(t,k,m)&=
 \exp\!\left(-\frac{\|\mathbf c_{t,k}-\widehat{\mathbf c}_{t,m}\|_2^2}{2\sigma^2}\right),
\end{align}
where $\sigma=0.1$ and clipping acts coordinate-wise.

Let $v_{t,k}=\mathbf1[p^{\mathrm{vis}}_{t,k}>0.5]$ and let $a_{t,k}$ indicate an accepted association. For an accepted pair, $m^\star$ is its selected location and $\bar h_{m^\star}$ is the state stored at $m^\star$ when the pair is processed. With $\beta=0.9$, initialize $h^{\mathrm{mem}}_{1,k}=h_{1,k}$. For $t>1$, the state returned to the tokenizer is
\begin{equation}
 h^{\mathrm{mem}}_{t,k}=\begin{cases}
 \beta\bar h_{m^\star}+(1-\beta)h_{t,k}, & a_{t,k}=1,\\
 h_{t,k}, & v_{t,k}=1,\ a_{t,k}=0,\\
 h^{\mathrm{mem}}_{t-1,k}, & v_{t,k}=0.
 \end{cases}
 \label{eq:persistent_state_update}
\end{equation}
Accepted pairs are processed in descending score order, and the one-to-one constraint applies to memory locations. Matched entries are read from a snapshot taken before any transfer, so slots that exchange entries in one frame keep distinct identifiers. At a visible location, the stored trajectory state is replaced by $h^{\mathrm{mem}}_{t,k}$; inactive locations keep their state. Slot and appearance prototypes follow $\bar y\leftarrow\beta\bar y+(1-\beta)y_{t,k}$ for $y\in\{S,x\}$ after any transfer. The two token-level event indicators are
\begin{equation}
 b_{t,k}=\bigl[a_{t,k},\ |v_{t,k}-v_{t-1,k}|\bigr],\qquad t>1,
 \label{eq:event_flags_exact}
\end{equation}
with $b_{1,k}=(0,0)$. At the first frame, identifiers are assigned to slots whose visibility gate is on. For $t>1$, a new identifier also requires the existence gate, and an observation enters the assignment map when its identifier is valid and both confidence gates exceed $0.5$. Algorithm~\ref{alg:memory} summarizes the entry updates and narrative construction.

\newcounter{memoryalgorithm}
\begin{table}[t]
\centering
\begin{minipage}{\columnwidth}
\refstepcounter{memoryalgorithm}\label{alg:memory}
\hrule\smallskip
\textbf{Algorithm \thememoryalgorithm: Persistent object narratives}\par
\smallskip
\textbf{Input:} slot observations, appearance, local state, existence/visibility predictions, and scene cuts.
\begin{enumerate}
\item Initialize memory and fresh IDs for visible first-frame slots.
\item For each later frame, retain assignments of continuing visible slots, except at a scene cut, and collect valid stored entries not claimed by them.
\item Score newly visible slots, and every visible slot at a scene cut, against candidate entries with the five cues; use appearance cues at scene cuts and reset local motion.
\item Greedily accept one-to-one pairs above the threshold. Transfer matched IDs and memory, and invalidate each distinct source entry.
\item Give unmatched visible, existing slots fresh IDs. Update visible memory summaries and hold invisible states.
\item Clear the assignments of slots whose existence gate is off; give a fresh ID to any visible, existing slot that still lacks one. Record valid visible assignments in $\pi_{t,k}$.
\item After the last frame, group recorded observations by ID and time, retain at most $M$ narratives, and form identity/state tokens.
\end{enumerate}
\hrule
\end{minipage}
\end{table}

\subsection{Scene-Cut Handling}
\label{app:scene_cut}

Scene cuts are detected with HSV-histogram Bhattacharyya distance using a threshold of $0.35$. The same cut mask is used throughout the pipeline:
\begin{itemize}
\item recurrent slot initialization is reset at the first frame of each shot;
\item cross-shot patch displacement is zeroed;
\item trajectory differences and causal-convolution history restart;
\item future-prediction targets crossing a cut are removed; and
\item every visible slot is matched again at the cut, using only slot-feature and pooled-appearance cues, and velocity is reset.
\end{itemize}
At a cut, every visible slot releases its identifier and is matched again using appearance cues, and local motion is reset.
\section{Typed Object-State Tokenization}
\label{app:typed_interface}

Typed object-state tokenization converts persistent narratives into the visual sequence of the language model through inventory selection, narrative pooling, and learned projection (Figures~\ref{fig:framework} and~\ref{fig:typed_interface}). Each inventory entry is an object reference: its identity token supplies an appearance reference across observations, and its state tokens describe evidence within individual segments. Shared object embeddings link the two token types, and segment embeddings specify when each state occurs.

\subsection{Object Inventory Selection}
\label{app:objectness}

The memory map may contain more persistent identifiers than the fixed object budget $M$. To reduce the effect of diffuse slot fragments, we define a soft objectness score from the slot-competitive area $\mu_{t,k}$ defined in Appendix~\ref{app:feature_attention} and the normalized patch entropy $H_{t,k}$. Let $s_{t,k}=\mu_{t,k}(1-H_{t,k})$, and let $s_{\min}$ and $s_{\max}$ be the minimum and maximum over all $(t,k)$ in the same clip. We use
\begin{equation}
  \eta_{t,k}=
  \frac{s_{t,k}-s_{\min}}
       {\max(s_{\max}-s_{\min},\varepsilon_\eta)},
  \qquad \varepsilon_\eta=10^{-6}.
\end{equation}
Thus $\eta_{t,k}\in[0,1]$ measures objectness relative to the other slots in the clip. During feature aggregation, we use $\widetilde\eta_{t,k}=\max(\eta_{t,k},0.1)$ and $\alpha_{t,k}=p^{\mathrm{vis}}_{t,k}\widetilde\eta_{t,k}$, where $p^{\mathrm{vis}}_{t,k}$ is the observation-confidence score described in Appendix~\ref{app:representation_memory}. Let $\mathcal I_{\mathrm{obs}}=\{v\geq0:\exists(t,k),\ \pi_{t,k}=v\}$ and $U=|\mathcal I_{\mathrm{obs}}|$. For each candidate identifier $v\in\mathcal I_{\mathrm{obs}}$, let $\mathcal{O}_v=\{(t,k):\pi_{t,k}=v\}$ be its assigned observations and $\mathcal{F}_v=\{t:\exists k,\,(t,k)\in\mathcal{O}_v\}$ its observed frames. We set $n_v^{\mathrm{vis}}=|\mathcal{F}_v|$. Let $\widehat h_{t,v}$ be the confidence--objectness weighted persistent trajectory state pooled over all slots assigned to $v$ in frame $t$. Let $\mathcal A_v=\{t\in\{1,\ldots,T-1\}:t,t+1\in\mathcal F_v\}$ contain adjacent observed-frame pairs. The state-change term is
\begin{equation}
  \overline{\Delta h}_v=
  \frac{\sum_{t\in\mathcal A_v}
  \|\widehat h_{t+1,v}-\widehat h_{t,v}\|_2}
  {\max(|\mathcal A_v|,1)}.
\end{equation}
We also define $\bar\eta_v$ as the mean $\eta_{t,k}$ over $\mathcal{O}_v$, floored at $0.1$. Its inventory score is
\begin{equation}
  \mathrm{Imp}_v=
  \bigl(n_v^{\mathrm{vis}}+0.5\,\overline{\Delta h}_v\bigr)\bar\eta_v.
\end{equation}
Let $R=\min(U,M)$ and $\mathcal U=\{1,\ldots,R\}$. If $U\leq M$, all observed identifiers are retained in increasing identifier order. If more than $M$ identifiers are present, the $M$ objects with largest $\mathrm{Imp}_v$ are retained and remapped in the selected order to $u\in\mathcal U$; as in the main paper, we reuse $\pi$ for the remapped association. This happens in at most 1.5\% of test videos (Appendix~\ref{app:representation_memory}). Discarded observations receive assignment $-1$. The remaining positions are padded and masked. The same inventory score and coefficient are used across datasets and budget settings.

\subsection{Identity and State Construction}
\label{app:identity_state_features}

For slot $k$ in frame $t$, its patch-normalized mask pools the raw SigLIP features into
\begin{equation}
  x_{t,k}=\sum_{p=1}^{P}A_{t,k,p}X_{t,p}\in\mathbb R^{D_v}.
\end{equation}
The identity summary of retained entry $u$ is the confidence- and objectness-weighted average
\begin{equation}
  r_u=
  \frac{\sum_{t,k:\,\pi_{t,k}=u}\alpha_{t,k}x_{t,k}}
       {\max\!\left(\sum_{t,k:\,\pi_{t,k}=u}\alpha_{t,k},\varepsilon\right)},
  \qquad \varepsilon=10^{-8}.
\end{equation}
For each assigned observation, let $b_{t,k}\in\{0,1\}^{2}$ mark an accepted association and a change in the gate $\mathbf{1}[p^{\mathrm{vis}}_{t,k}>0.5]$. Let $h^{\mathrm{mem}}_{t,k}$ be the persistent trajectory state returned by memory from the local trajectory sequence $h_{t,k}$. The state feature is
\begin{equation}
  g_{t,k}=[h^{\mathrm{mem}}_{t,k};o_{t,k};p^{\mathrm{vis}}_{t,k};b_{t,k};x_{t,k}]
  \in\mathbb R^{D_g}.
\end{equation}
Here, $D_g=D_h+D_o+3+D_v=1244$, and $[\,\cdot\,;\,\cdot\,]$ denotes concatenation. Thus, the state includes trajectory, geometry and mask statistics, observation confidence, two event indicators, and pooled appearance.

For $1\leq J\leq T$, set $d_j=\operatorname{round}(jT/J)$ for $j=0,\ldots,J$, with nearest-even rounding at ties. The uniform segment is $\mathcal I_j=\{d_{j-1}+1,\ldots,d_j\}$. Define $\Omega_{t,u}=\{k:\pi_{t,k}=u\}$ as the slots assigned to entry $u$ in frame $t$. For a per-slot feature $y_{t,k}$, its frame-level aggregate is
\begin{equation}
 \widehat y_{t,u}=
 \frac{\sum_{k\in\Omega_{t,u}}\alpha_{t,k}y_{t,k}}
 {\max(\sum_{k\in\Omega_{t,u}}\alpha_{t,k},\varepsilon)}.
\end{equation}
For a single assigned slot, it equals that slot's feature. For $g_{t,k}$, let $\mathcal{T}_{u,j}=\{t\in\mathcal{I}_j:\Omega_{t,u}\neq\emptyset\}$ be the valid frames of entry $u$ in segment $j$. Segment pooling is
\begin{equation}
  \bar g_{u,j}=
  \frac{1}{|\mathcal{T}_{u,j}|}
  \sum_{t\in\mathcal{T}_{u,j}}\widehat g_{t,u}.
\end{equation}
An object absent from a segment receives a masked state position, so this average is evaluated only when $|\mathcal{T}_{u,j}|>0$. Let $D_L=3584$ be the Qwen2 embedding dimension. The exact typed projections are
\begin{equation}
  z^{\mathrm{id}}_u
  =f_{\mathrm{id}}(\mathrm{LN}_{\mathrm{id}}(r_u))
  +e_{\mathrm{id}}+e_u,
\end{equation}
\begin{equation}
  z^{\mathrm{st}}_{u,j}
  =f_{\mathrm{st}}(\mathrm{LN}_{\mathrm{st}}(\bar g_{u,j}))
  +e_{\mathrm{st}}+e_u+e_j.
\end{equation}
Here, $f_{\mathrm{id}}:\mathbb R^{D_v}\rightarrow\mathbb R^{D_L}$ and $f_{\mathrm{st}}:\mathbb R^{D_g}\rightarrow\mathbb R^{D_L}$ are separate two-layer MLPs with 3584-dimensional hidden layers and GELU activations. $\mathrm{LN}_{\mathrm{id}}$ and $\mathrm{LN}_{\mathrm{st}}$ are separate layer normalizations. The embeddings $e_{\mathrm{id}}$, $e_{\mathrm{st}}$, $e_u$, and $e_j$ encode token type, object index, and segment index in $\mathbb R^{D_L}$. The projectors, layer normalizations, and embedding tables are the only modules trained in Stage III (34.39M parameters).

\paragraph{Empty inventories and masked positions.}
If no identifier is observed ($U=0$), which does not occur on the three test sets, the slot with the largest summed confidence forms one fallback entry over all frames (ties use the first slot):

\begin{equation}
 \begin{split}
 k^\star&=\arg\max_k\sum_{t=1}^{T}p^{\mathrm{vis}}_{t,k},\\
 r_1&=\frac1T\sum_{t=1}^{T}x_{t,k^\star},\qquad
 \widehat g_{t,1}=g_{t,k^\star}.
 \end{split}
 \label{eq:empty_inventory_fallback}
\end{equation}
For this fallback, use $R=1$, $\mathcal U=\{1\}$, and $\mathcal T_{1,j}=\mathcal I_j$. Unused entry positions and unobserved segment positions stay zero in the visual sequence, skip projection, and are masked. Identity and state attention masks are

\begin{equation}
 m^{\mathrm{id}}_u=\mathbf1[u\in\mathcal U],\qquad
 m^{\mathrm{st}}_{u,j}=\mathbf1[u\in\mathcal U,\ |\mathcal T_{u,j}|>0].
 \label{eq:visual_attention_masks}
\end{equation}
Masked positions are excluded as keys in the LLM attention. The valid identity and state positions are indexed by $\mathcal U$ and $\mathcal V=\{(u,j):m^{\mathrm{st}}_{u,j}=1\}$.

\subsection{Serialization and Attention Masking}
\label{app:serialization}

With object capacity $M$ and $J$ temporal segments, the visual sequence contains $M+MJ$ positions. The default $M=16,J=8$ configuration gives
\begin{equation}
  M(1+J)=16+16\times8=144
\end{equation}
visual tokens, ordered as $M$ identity positions followed by $J$ segment-major blocks of $M$ state positions. The projectors preserve these positions while mapping their features to the LLM embedding dimension. Unused entries and unobserved object--segment pairs are padded and masked. The reported token count includes all allocated positions, giving the same sequence length for every video.

\section{Training and Evaluation Details}
\label{app:training}

\paragraph{Training data.}
We train on VideoInstruct-100K~\cite{maaz2024video}, which pairs ActivityNet videos with approximately 100K instruction conversations. Stages I--II learn visual grouping and local-state features from the videos without segmentation or QA labels. Stage III uses the paired conversations to align the object-state tokens with the language model. Each completed visual stage is frozen before training the next stage.

\paragraph{Evaluation data.}
We report benchmark comparisons and component analyses on the official test sets of MSVD-QA, MSRVTT-QA~\cite{xu2017video}, and ActivityNet-QA~\cite{yu2019activitynet}.

\subsection{Training Overview}

The three rows of Table~\ref{tab:supp_training} correspond to the dashed training regions in Figure~\ref{fig:framework}: object grouping, local-state learning, and language alignment. Only the modules listed for each stage are trained; the reconstruction decoder and future-prediction heads provide auxiliary visual supervision.

\begin{table*}[t]
  \centering
\caption{Training configuration. SRL denotes Synergistic Representation Learning. Modules learned in an earlier stage are frozen before the next stage.}
  \label{tab:supp_training}
  \begin{tabular}{@{}p{0.07\textwidth}p{0.23\textwidth}p{0.27\textwidth}p{0.35\textwidth}@{}}
    \toprule
    Stage & Data and sampling & Trainable modules & Optimization and objectives \\
    \midrule
    I & ActivityNet videos; 100 sampled frames; preferentially within-shot
    8-frame clips; $384\times384$ input & Feature adapter, recurrent slot
    encoder, slot predictor, feature decoder & 50K steps; AdamW,
    $1.5\times10^{-4}$; feature reconstruction, scheduled SlotContrast, and
    scheduled SRL \\
    II & Same unlabeled videos; full sampled sequence & Trajectory encoder,
    two future-prediction heads, presence estimator & 20 epochs; Adam,
    $2\times10^{-4}$; trajectory-prediction weight $0.5$ and presence weight $0.2$ on $\mathcal{L}_{\mathrm{exist}}+0.5\mathcal{L}_{\mathrm{visible}}$ \\
    III & VideoInstruct-100K conversations; same video sampling &
    Identity and state projectors, layer normalizations, and typed embeddings & 3
    epochs; projector learning rate $2\times10^{-5}$; answer-only
    autoregressive cross-entropy \\
    \bottomrule
  \end{tabular}
\end{table*}

\subsection{Stage I: Object Slot Representation}
\label{app:stage1_training}

At optimization step $n$, the Stage I objective is
\begin{equation}
  \mathcal{L}^{(1)}(n)=\mathcal{L}_{\mathrm{rec}}
  +\lambda_{\mathrm{SC}}(n)\mathcal{L}_{\mathrm{SC}}
  +\lambda_{\mathrm{SRL}}(n)\mathcal{L}_{\mathrm{SRL}}.
\end{equation}
$\mathcal{L}_{\mathrm{rec}}$ is mean-squared reconstruction of frozen SigLIP patch features. $\mathcal{L}_{\mathrm{SC}}$ is the SlotContrast objective over adjacent frames in the same shot. $\mathcal{L}_{\mathrm{SRL}}$ denotes the scheduled SRL component: early redundancy regularization or later encoder/decoder consistency. The functions $\lambda_{\mathrm{SC}}(n)$ and $\lambda_{\mathrm{SRL}}(n)$ are their step-dependent weights and are zero when the corresponding component is inactive.

Feature reconstruction has unit weight throughout training. SlotContrast has target weight $0.05$, starts at step 1,500, and uses a 1,500-step linear warm-up. Redundancy regularization is active during the first 5,000 steps with phase weight $0.3$ and redundancy-loss weight $0.05$. Encoder/decoder consistency starts at step 10,000 with a 1,000-step warm-up and target weight $0.05$. Temporal pairs crossing scene cuts are removed from all temporal objectives. The Stage I representation is then frozen for the downstream stages.

\subsection{Stage II: Trajectory and Presence}
\label{app:stage2_training}

Let $f_\delta:\mathbb R^{D_h}\rightarrow\mathbb R^{D_o}$ be the prediction head for temporal offset $\delta\in\{1,3\}$, and let $\mathcal{V}_\delta$ contain minibatch source-slot triples $(b,t,k)$ for which $t+\delta\leq T$ and the interval from $t$ to $t+\delta$ does not cross a scene cut. We use the same future slot index because recurrent initialization provides local slot-index continuity within a shot; persistent cross-slot association is introduced only after the trajectory encoder. We use the reliability $\xi_{b,t,k}=\mu_{b,t,k}(1-H_{b,t,k})$ and optimize
\begin{equation}
  \mathcal{L}^{\delta}_{\mathrm{pred}}=
  \frac{\sum_{(b,t,k)\in\mathcal{V}_\delta}\xi_{b,t,k}
  \left\|f_\delta(h_{b,t,k})-\mathrm{sg}(o_{b,t+\delta,k})\right\|_2^2}
  {D_o\max\!\left(\sum_{(b,t,k)\in\mathcal{V}_\delta}\xi_{b,t,k},
  N_\delta\varepsilon\right)},
\end{equation}
where $D_o=25$, $N_\delta=B(T-\delta)K$ is the number of source-slot triples before cut masking for batch size $B$, and $\varepsilon=10^{-8}$. The $N_\delta\varepsilon$ term corresponds to normalizing the masked weights by $\max(\operatorname{mean}(\xi),\varepsilon)$; an all-masked term is zero. Here $\|\cdot\|_2$ is the Euclidean norm, and $\mathrm{sg}$ stops gradients through the target observation. For presence supervision, the confidence $c_{t,k}$ from Appendix~\ref{app:trajectory_state} gives $y^e_{t,k}=\mathbf{1}[c_{t,k}>3]$ and $y^v_{t,k}=\mathbf{1}[c_{t,k}>5]$. Let $\operatorname{BCE}(p,y)=-y\log p-(1-y)\log(1-p)$, evaluated stably from logits, and let $\mathcal L_{\mathrm{exist}}$ and $\mathcal L_{\mathrm{visible}}$ average it over all $BTK$ entries for their respective targets. Then the complete Stage II objective is
\begin{equation}
\begin{aligned}
  \mathcal{L}^{(2)}&=0.5\,\mathcal{L}_{\mathrm{pred}}+0.2\,\mathcal{L}_{\mathrm{pres}},\\
  \mathcal{L}_{\mathrm{pred}}&=\frac{1}{2}\sum_{\delta\in\{1,3\}}\mathcal{L}^{\delta}_{\mathrm{pred}},\\
  \mathcal{L}_{\mathrm{pres}}&=\mathcal{L}_{\mathrm{exist}}+0.5\,\mathcal{L}_{\mathrm{visible}}.
\end{aligned}
\end{equation}
The prediction loss learns short-range state changes, while the attention-derived targets supervise the confidence gates. The trained modules feed the fixed persistent-memory rule described in Appendix~\ref{app:persistent_memory}.

\subsection{Stage III: Language Alignment}
\label{app:stage3_training}

Stage III uses the VideoInstruct-100K instruction conversations and the freeze policy summarized in Table~\ref{tab:supp_training}. Training uses batch size 8, four-step gradient accumulation, cosine decay, and a 3\% warm-up ratio. The standard Qwen2 chat template places the serialized object-state sequence before the question, and the loss is applied only to assistant-answer tokens. Let $\mathcal{Y}$ denote the set of answer-token positions, $a_\ell$ the answer token at position $\ell$, $a_{<\ell}$ its preceding answer prefix, $q$ the question, $\mathcal{Z}(V)$ the visual-token sequence, and $p_{\mathrm{LLM}}$ the next-token distribution. The objective is
\begin{equation}
  \mathcal{L}^{(3)}=-\frac{1}{|\mathcal{Y}|}
  \sum_{\ell\in\mathcal{Y}}
  \log p_{\mathrm{LLM}}
  \bigl(a_\ell\mid \mathcal{Z}(V),q,a_{<\ell}\bigr).
\end{equation}

Gradients pass through the frozen language model and update the identity/state projectors, layer normalizations, and type/object/segment embeddings. The upstream visual modules remain frozen, so language alignment operates on fixed object narratives.

\subsection{Implementation Details}
\label{app:reproducibility_settings}

\paragraph{Training environment.}
Training and VideoQA evaluation of \ours{} use two NVIDIA A800 GPUs. Trainable stages use bfloat16 mixed precision, with random seed 42 for training and data loading. Stage III uses batch size 8 per GPU and four-step gradient accumulation, giving an effective global batch size of 64.

\paragraph{Answer generation and evaluation.}
For \ours{} and its ablations, open-ended answers are decoded deterministically with sampling disabled, the tokenizer end-of-sequence token as the stopping condition, and at most 64 new tokens. Following Video-ChatGPT, we use gpt-3.5-turbo with temperature zero and a 64-token response limit. The prompt asks the judge to compare the prediction with the reference, accept synonymous or paraphrased answers, and return binary correctness together with an integer quality score from 0 to 5. Malformed responses are retried up to three times. The same evaluator is used for all model variants. Appendix~\ref{app:judge_prompt} gives the complete prompt.

\paragraph{Average-pooling baseline.}
The same-recipe baseline in the lower block of Table~\ref{tab:compact_comparison} replaces the object-state interface with average pooling on a $16\times3\times3$ spatiotemporal grid. The same 100 frames are divided into 16 consecutive temporal bins, and each $27\times27$ SigLIP patch grid into $3\times3$ cells of $9\times9$ patches. Averaging the features within each bin and cell yields $16\times3\times3=144$ visual tokens, all of them active. All other components and settings, including the projector design, the frozen Qwen2-7B-Instruct, the Stage III data and schedule, and the evaluation protocol, follow \ours{}. This baseline is evaluated on ActivityNet-QA.

\subsection{Evaluation Prompt}
\label{app:judge_prompt}

The evaluator receives the following system instruction and question--answer template.

\begin{quote}
\small
\textbf{System.} You are an intelligent chatbot designed for evaluating the correctness of generative outputs for question-answer pairs. Your task is to compare the predicted answer with the correct answer and determine if they match meaningfully. Here's how you can accomplish the task:
------
\#\#INSTRUCTIONS: 

- Focus on the meaningful match between the predicted answer and the correct answer.

- Consider synonyms or paraphrases as valid matches.

- Evaluate the correctness of the prediction compared to the answer.
\end{quote}

\begin{quote}
\small
\textbf{User.} Please evaluate the following video-based question-answer pair:

Question: \{question\}
Correct Answer: \{reference answer\}
Predicted Answer: \{predicted answer\}

Provide your evaluation only as a yes/no and score where the score is an integer value between 0 and 5, with 5 indicating the highest meaningful match. Please generate the response in the form of a Python dictionary string with keys 'pred' and 'score', where value of 'pred' is a string of 'yes' or 'no' and value of 'score' is in INTEGER, not STRING.
DO NOT PROVIDE ANY OTHER OUTPUT TEXT OR EXPLANATION. Only provide the Python dictionary string. For example, your response should look like this: \{'pred': 'yes', 'score': 4\}.
\end{quote}

\FloatBarrier

\section{Additional Experiments and Analysis}
\label{app:additional_results}

This appendix reports the SlotContrast ablation, an analysis of trajectory states, question-type results, and failure cases. All QA results use the settings in Appendix~\ref{app:reproducibility_settings}.

\subsection{Temporal Consistency of Slots}
\label{app:component_ablations}
\label{app:diagnostics}

Removing SlotContrast from Stage I lowers ActivityNet-QA accuracy from 55.2\% to 50.7\% with the same 144-token interface (Table~\ref{tab:supp_slotcontrast}). Temporally consistent slots therefore matter for the narratives built on them. Appendix~\ref{app:direct_evaluation} evaluates the slot groups directly against reference masks.

\begin{table}[t]
\centering
\caption{Effect of SlotContrast on ActivityNet-QA. $\Delta$ is the accuracy change from the full model.}
\label{tab:supp_slotcontrast}
{\small
\setlength{\tabcolsep}{5pt}
\begin{tabular}{@{}lrrr@{}}
\toprule
Variant & \tok{} & Acc. & $\Delta$ \\
\midrule
Full model & 144 & 55.2 & -- \\
w/o SlotContrast & 144 & 50.7 & -4.5 \\
\bottomrule
\end{tabular}}
\end{table}

\subsection{Predictive Information in Local States}
\label{app:state_prediction}

We test whether the learned trajectory state $h_{t,k}$ carries information about future observations. The analysis uses the fifty ActivityNet and ninety DAVIS videos of Appendix~\ref{app:direct_evaluation}, split by video into two thirds for fitting and one third for testing (seed 0). For $\delta\in\{1,3\}$, pairs keep the same raw slot index, require $p^{\mathrm{vis}}>0.5$ at both endpoints, and exclude scene boundaries from $t-2$ through $t+\delta$.

\begin{figure}[t]
\centering
 \includegraphics[width=\columnwidth]{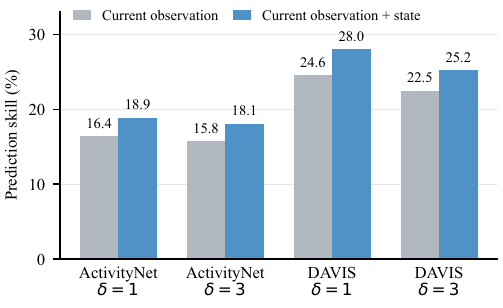}
 \caption{\textbf{Local-state prediction.} Future-observation prediction skill from the current observation alone and with the learned trajectory state. Skill is the percentage reduction in standardized squared error relative to copying the current observation.}
 \label{fig:supp_representation_diagnostics}
\end{figure}

We fit ridge-regression predictors (linear models with $L_2$ regularization $10^{-2}$) to the residual $o_{t+\delta,k}-o_{t,k}$. The inputs are the current observation $o_{t,k}$ or the observation with the learned state $[o_{t,k};h_{t,k}]$. Input means and scales, and the target scale $D_\sigma=\operatorname{diag}(\operatorname{std}_{\mathrm{train}}(o_{t+\delta,k})+10^{-6})$, are estimated on fitting videos only. Let $f_{t,k}$ be a predictor's reconstructed future observation and $\mathcal E_\delta$ the set of valid held-out pairs, with video indices suppressed. We measure

\begin{align}
 E_\delta(f)&=\sum_{(t,k)\in\mathcal E_\delta}
 \|D_\sigma^{-1}(f_{t,k}-o_{t+\delta,k})\|_2^2,\\
 \operatorname{Skill}_\delta(f)&=100\left(1-\frac{E_\delta(f)}{E_\delta(f^{\mathrm{copy}})}\right),
 \label{eq:prediction_skill}
\end{align}
where $f^{\mathrm{copy}}_{t,k}=o_{t,k}$. Skill is the percentage reduction in prediction error relative to copying the current observation, which has zero skill.

Adding the learned state raises skill in all four dataset and offset groups (Figure~\ref{fig:supp_representation_diagnostics}). At $\delta=1$, skill rises from 16.4\% to 18.9\% on ActivityNet and from 24.6\% to 28.0\% on DAVIS. The state thus carries predictive context beyond the current observation, which state tokens pass to the language model. The held-out sets contain 3,224/2,654 pairs on ActivityNet and 7,639/7,033 on DAVIS for $\delta=1/3$.

\subsection{Question Types}
\label{app:question_types}

Table~\ref{tab:supp_activitynet_type} reports ActivityNet-QA results by the released answer-type labels, which partition the 8,000 test questions. Temporal and spatial questions, which relate two events or two objects, are the hardest types.

\begin{table}[t]
  \centering
\caption{ActivityNet-QA accuracy (\%) by answer type.}
  \label{tab:supp_activitynet_type}
  \setlength{\tabcolsep}{5pt}
  {\small
  \begin{tabular}{@{}lrr@{}}
    \toprule
    Type & \#Q & Acc. \\
    \midrule
    Overall & 8,000 & 55.2 \\
    \midrule
    Yes/No & 2,094 & 76.0 \\
    Other & 1,499 & 56.6 \\
    Temporal & 800 & 31.3 \\
    Motion & 800 & 55.4 \\
    Spatial & 800 & 30.8 \\
    Color & 697 & 48.0 \\
    Number & 606 & 47.4 \\
    Location & 386 & 72.8 \\
    Object & 318 & 43.1 \\
    \bottomrule
  \end{tabular}}
\end{table}

\subsection{Failure Cases}
\label{app:failure_cases}

\begin{figure*}[t]
\centering
  \includegraphics[width=\textwidth]{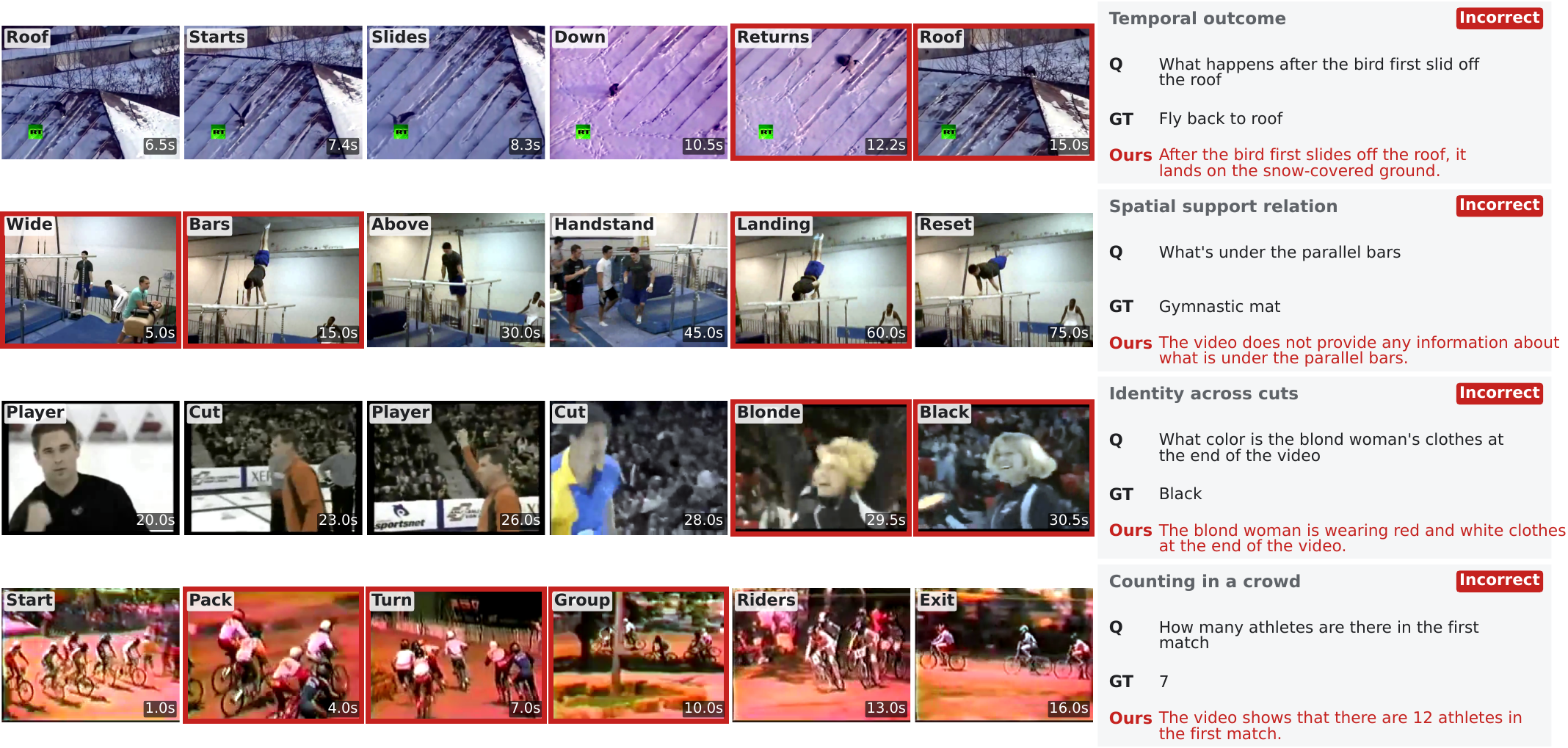}
  \caption{Representative ActivityNet-QA failures. Each row contains six event-aligned source-video frames. Red borders mark evidence that conflicts with the prediction. From top to bottom, the cases involve a temporal outcome, spatial support, target appearance after cuts, and counting in a crowded scene.}
  \label{fig:supp_failure_case}
\end{figure*}

Figure~\ref{fig:supp_failure_case} shows four answer errors on ActivityNet-QA: the bird returns to the roof rather than landing on snowy ground; a mat is visible beneath the gymnast; the target woman wears black at the end; and the cyclist count is seven rather than twelve. Red borders mark the evidence that each prediction misses.

\section{Object Grouping and Association Analysis}
\label{app:direct_evaluation}

\subsection{Evaluation Setup}
\label{app:direct_protocol}

We evaluate the visual representations learned in Stages I--II with their parameters fixed: the spatial groups that collect observations and the links that connect them across time.

Grouping is evaluated on fifty ActivityNet-QA test videos and on ninety DAVIS-2017 training and validation videos with human instance masks~\cite{ponttuset2017davis}; neither set is used for visual training. Of the fifty ActivityNet videos, ten calibrate the appearance-only baseline and forty are used for association. For ActivityNet, Qwen3-VL-8B-Instruct~\cite{qwen3vl8b} proposes object noun phrases and SAM~3~\cite{carion2025sam3} tracks the corresponding instances as pseudo-references. We build these references at twice the model's frame-sampling density. A second reference built at the model's own frame rate agrees on 96.8\% of identity links.

We report percentile 95\% confidence intervals (CIs) from 2,000 video-level bootstrap samples. Grouping scores are averaged over videos. Association ratios pool numerators and denominators across videos, and resamples with a zero denominator count as zero.

\subsection{Spatial and Within-Shot Grouping}
\label{app:grouping_diagnostics}

The competitive slot attention is converted to a hard partition by taking the largest slot weight at each of the $27\times27$ patch cells. A reference cell receives the object covering the greatest area, including background. Frame foreground adjusted Rand index (FG-ARI) measures chance-corrected agreement between partitions of foreground cells: zero corresponds to chance and one to identical grouping. Mean best overlap (mBO) averages each reference object's largest intersection-over-union with any predicted group.

The feature baseline uses $K=11$ clusters on L2-normalized SigLIP patch features, with 25 $k$-means iterations and a fixed initialization seed. For temporal metrics, consecutive-frame clusters are matched by Hungarian assignment using centroid cosine similarity. Fixed spatial groups partition the coordinate grid into eleven regions. Video FG-ARI and video mBO concatenate the frames of each shot with group indices kept across time, so they also test within-shot continuity.

\begin{table*}[t]
\centering
\caption{Spatial and within-shot grouping. ActivityNet uses fifty videos with pseudo-reference masks; DAVIS uses ninety videos with human masks. Each method has eleven groups.}
\label{tab:supp_grouping_complete}
\setlength{\tabcolsep}{4pt}
{\small
\begin{tabular}{@{}llrrrr@{}}
\toprule
Reference & Grouping & Frame FG-ARI [95\% CI] & mBO & Video FG-ARI [95\% CI] & Video mBO \\
\midrule
ActivityNet & Fixed groups & 0.210 [0.176, 0.244] & 0.203 & 0.149 [0.125, 0.177] & 0.152 \\
ActivityNet & Feature k-means & 0.152 [0.127, 0.180] & 0.213 & 0.081 [0.065, 0.099] & 0.130 \\
ActivityNet & Learned slots & 0.306 [0.258, 0.358] & 0.271 & 0.222 [0.187, 0.263] & 0.196 \\
\midrule
DAVIS & Fixed groups & 0.140 [0.104, 0.179] & 0.223 & 0.061 [0.042, 0.081] & 0.169 \\
DAVIS & Feature k-means & 0.120 [0.094, 0.151] & 0.232 & 0.022 [0.015, 0.029] & 0.114 \\
DAVIS & Learned slots & 0.270 [0.214, 0.328] & 0.329 & 0.135 [0.097, 0.177] & 0.260 \\
\bottomrule
\end{tabular}
}
\end{table*}

With the same eleven-group budget, learned slots outperform fixed spatial groups and feature clustering on all four measures in both datasets (Table~\ref{tab:supp_grouping_complete}).

\subsection{Association Against Pseudo-Reference Tracks}
\label{app:conditional_association}

\paragraph{Matched observations.}
A slot represents a reference object when at least half of its assigned patch area lies on that object and it covers at least 10\% of the object's area. We keep matched observations that also pass both confidence gates, giving a common set $\mathcal O$ for all assignment variants. These observations cover 6.0\% [4.5, 8.1] of pseudo-reference object-frames. Half of these object-frames show objects smaller than 1\% of the frame, which no slot represents; coverage is 12.4\% for objects covering at least 1\% of the frame and 19.6\% for objects covering at least 5\%.

\paragraph{Assignments.}
The \emph{Recurrent slots (no memory)} baseline uses the raw indices of the learned Stage I slots as entry identifiers. \emph{Appearance-only} association matches on pooled SigLIP cosine similarity, with threshold .50 chosen on the ten calibration videos. \emph{Full memory} uses its default threshold of .30. All variants use the same observations. The two memory-based variants maintain their own sequential memory states.

\paragraph{Identity consistency.}
We adapt identity F1 (IDF1)~\cite{ristani2016performance} to $\mathcal O$. For each video with $|\mathcal O|>0$, let $C_{i,u}$ count observations with model identity $i$ and reference object $u$. The score is
\begin{equation}
 \mathrm{IDF1}_{\mathcal O}
 =\frac{\max_{\sigma}\sum_i C_{i,\sigma(i)}}{|\mathcal O|},
 \label{eq:conditional_idf1}
\end{equation}
where $\sigma$ is a one-to-one matching between the two identity sets, with unmatched identities contributing zero. Because each observation in $\mathcal O$ has exactly one predicted and one reference identity, this equals the standard IDF1 on $\mathcal O$. Dataset scores sum the matched counts and observation counts over videos.

\paragraph{Pairwise links.}
We use the pair-counting formulation of precision and recall~\cite{amigo2009comparison}. Within each video, let $\mathcal P$ contain pairs of observations from $\mathcal O$ at different frames, $G\subseteq\mathcal P$ the pairs sharing a reference object, and $H\subseteq\mathcal P$ the pairs sharing a model identity. Then
\begin{equation}
 \mathrm{Pair\ P}=\frac{|G\cap H|}{|H|},\qquad
 \mathrm{Pair\ R}=\frac{|G\cap H|}{|G|}.
 \label{eq:pairwise_metrics}
\end{equation}
Pair P measures the purity of predicted links, and Pair R the fraction of same-object pairs that are linked. Let $X\subseteq\mathcal P$ contain the pairs whose raw slot indices differ. Cross-slot metrics apply the same definitions to this subset:
\begin{equation}
 \mathrm{Cross\ P}=\frac{|G\cap H\cap X|}{|H\cap X|},\qquad
 \mathrm{Cross\ R}=\frac{|G\cap H\cap X|}{|G\cap X|}.
 \label{eq:cross_slot_metrics}
\end{equation}
A reference gap occurs when the reference mask of the earlier observation's object is absent from at least one intermediate sampled frame. With $Q\subseteq\mathcal P$ the pairs spanning such a gap, $\mathrm{Gap\ R}=|G\cap H\cap Q|/|G\cap Q|$. Restricting to pairs from the same raw slot, $Y=\mathcal P\setminus X$, gives
\begin{equation}
 \mathrm{Same\text{-}slot\ gap\ R}
 =\frac{|G\cap H\cap Q\cap Y|}{|G\cap Q\cap Y|},
 \label{eq:same_slot_gap_recall}
\end{equation}
and same-slot precision $|G\cap H\cap Y|/|H\cap Y|$. Pair counts are summed across videos before each ratio is computed.

\begin{table}[t]
\centering
\caption{\textbf{Links within and across raw slots} (dense pseudo-reference; \%). Gap R counts same-object pairs that span a reference gap. $\dagger$: fixed by assigning recurrent-slot indices as identities; --: zero denominator.}
\label{tab:supp_assoc_breakdown}
\setlength{\tabcolsep}{3.8pt}
{\small
\begin{tabular}{@{}lrrrr@{}}
\toprule
& \multicolumn{2}{c}{Same raw slot} & \multicolumn{2}{c}{Across raw slots} \\
\cmidrule(lr){2-3}\cmidrule(l){4-5}
Assignment & Precision & Gap R & Recall & Precision \\
\midrule
\shortstack[l]{Recurrent slots\\(no memory)} & 93.6 & 100.0$^{\dagger}$ & 0.0$^{\dagger}$ & -- \\
Appearance-only & 98.0 & 32.0 & 8.8 & 38.3 \\
Full memory & 96.7 & 95.0 & 15.0 & 72.0 \\
\bottomrule
\end{tabular}}
\end{table}

\begin{table*}[t]
\centering
\caption{\textbf{Object association and matching design.} Metrics use the matched observations and dense pseudo-reference of Appendix~\ref{app:conditional_association}. All values are percentages; IDF1$_{\mathcal O}$ includes a 95\% video-bootstrap interval. Cue removals change only the matching score.}
  \label{tab:supp_assoc_design}
{\small
\setlength{\tabcolsep}{4pt}
\begin{tabular}{@{}lrrrrrr@{}}
\toprule
Assignment & IDF1$_{\mathcal O}$ [95\% CI] & Pair P & Pair R & Gap R & Cross R & Cross P \\
\midrule
Recurrent slots (no memory) & 64.4 [55.1, 73.1] & 93.6 & 61.5 & 45.7 & 0.0 & -- \\
Appearance-only & 56.2 [48.5, 64.5] & 88.2 & 47.7 & 18.5 & 8.8 & 38.3 \\
Full memory & 67.0 [57.6, 76.0] & 93.8 & 65.0 & 51.8 & 15.0 & 72.0 \\
\midrule
w/o slot features & 61.4 [52.0, 70.4] & 92.0 & 51.5 & 22.0 & 14.2 & 66.0 \\
w/o pooled appearance & 65.4 [56.0, 74.4] & 95.3 & 61.2 & 46.2 & 5.0 & 68.0 \\
w/o trajectory state & 66.6 [57.2, 75.6] & 93.7 & 64.7 & 51.3 & 14.6 & 71.5 \\
w/o recent observation & 66.7 [57.3, 75.7] & 93.7 & 64.8 & 51.5 & 14.8 & 71.2 \\
w/o position & 66.4 [57.0, 75.4] & 93.6 & 64.6 & 51.0 & 14.7 & 70.8 \\
Equal weights & 66.5 [57.1, 75.5] & 93.7 & 64.7 & 51.2 & 14.9 & 71.0 \\
Hungarian & 66.9 [57.5, 75.9] & 93.8 & 65.0 & 51.7 & 15.0 & 71.8 \\
\bottomrule
\end{tabular}}
\end{table*}

\begin{figure*}[t]
  \centering
  \includegraphics[width=\textwidth]{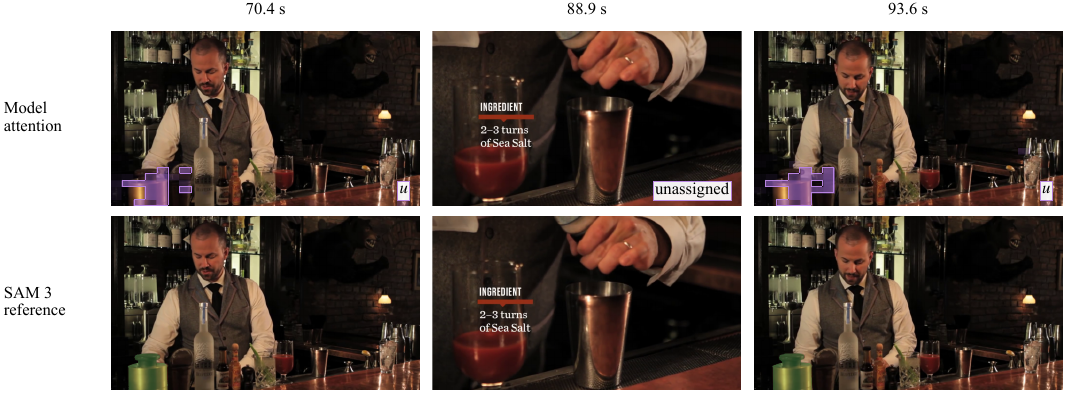}
  \caption{\textbf{Full-frame evidence for the cut-away example.} The target is the lidded metal container at the left edge of the wide shot, distinct from the central mixing cup in the close-up. The upper row shows the model's attention for entry $u$, and the lower row the SAM~3 pseudo-reference mask. The marked windows in Figure~\ref{fig:physical_reappearance} are display enlargements.}
  \label{fig:supp_cutaway_full}
\end{figure*}

\paragraph{Results.}
Recurrent slots alone reach IDF1$_{\mathcal O}$ 64.4\%, reflecting the temporal consistency learned in Stage I, and full memory raises it to 67.0\% (Table~\ref{tab:supp_assoc_design}).
Table~\ref{tab:supp_assoc_breakdown} separates links within and across raw slots. Within a raw slot, memory links are more precise than raw-index links (96.7\% vs.\ 93.6\%), because a returning slot that shows a different object can receive a different entry, and memory keeps 95.0\% of same-object pairs that span a reference gap, compared with 32.0\% for appearance-only matching. Across raw slots, which hold 54.3\% of the same-object pairs spanning a gap, memory links 15.0\% of same-object pairs at 72.0\% precision; raw indices form no such links.

\paragraph{Entry retention across a cut.}
Figure~\ref{fig:supp_cutaway_full} shows the example of Figure~\ref{fig:physical_reappearance} on complete frames at three moments. At 88.9\,s the camera shows a different mixing cup and the target container is out of view; its entry receives no observation and keeps its stored state. When the wide view returns, the container is assigned to the same entry.

\subsection{Matching Rule and Observation Confidence}
\label{app:association_design}

\paragraph{Matching cues.}
Table~\ref{tab:supp_assoc_design} also varies the matching rule on the same observations and gates. Each cue ablation removes that term and renormalizes the remaining weights. Slot-feature similarity matters most for identity consistency: removing it lowers IDF1$_{\mathcal O}$ from 67.0\% to 61.4\%. Pooled appearance drives links across raw slots: removing it lowers cross-slot recall from 15.0\% to 5.0\%. The rule is robust to the remaining choices: removing any single motion-related cue, using equal weights, or using Hungarian assignment changes IDF1$_{\mathcal O}$ by less than one point.

\paragraph{Matching threshold.}
Sweeping the acceptance threshold over $\{.20,.25,.30,.35,.40\}$ keeps IDF1$_{\mathcal O}$ between 66.2\% and 67.0\%. Raising it from .30 to .40 increases cross-slot precision from 72.0\% to 76.8\% and lowers cross-slot recall from 15.0\% to 11.8\%.

\paragraph{Confidence gates.}
The gates decide which observations enter narratives. To examine their thresholds, we replace the predicted gates with their binary training targets $\mathbf1[c>\tau_e]$ and $\mathbf1[c>\tau_v]$, keeping the visual features fixed. The default pair $(3,5)$ gives 6.2\% object-frame coverage. The looser pair $(2,4)$ raises coverage to 9.9\% at 92.7\% pair precision, and the stricter pair $(4,6)$ lowers it to 3.8\% at 96.8\%. The thresholds thus trade coverage for link precision.

\section{Inference Efficiency}
\label{app:runtime}

We measure the cost of building object narratives and answering with them. All measurements use one NVIDIA RTX~3090 with CUDA synchronization.

\subsection{Language-Model Context Cost}
\label{app:llm_profile}

To isolate visual context length, we run the frozen Qwen2-7B in bf16 with a fixed text prompt and a 32-token output. The visual input consists of 144, 576, 2,304, 6,272, or 19,600 random embeddings, all attended. We measure prefill latency and the average time per generated token after warm-up.

For $L$ visual positions, the visual key/value (KV) cache capacity is
\begin{equation}
 B_{\mathrm{KV}}=2L\,N_{\mathrm{layers}}N_{\mathrm{KVheads}}d_{\mathrm{head}}b,
\end{equation}
where the factor two accounts for keys and values, $N_{\mathrm{layers}}=28$, $N_{\mathrm{KVheads}}=4$, $d_{\mathrm{head}}=128$, and $b=2$ bytes for bf16. Thus 144 positions require 7.875 MiB. Longer visual sequences increase prefill latency and KV storage substantially, whereas per-token decoding changes little (Table~\ref{tab:supp_llm_scaling}). The default interface allocates 144 positions, and its KV capacity follows this allocated length.

\begin{table}[!htbp]
\centering
\caption{Effect of visual context length on Qwen2-7B. KV is the calculated visual-prefix capacity; peak GPU memory includes the resident models and intermediate activations.}
\label{tab:supp_llm_scaling}
\setlength{\tabcolsep}{3pt}
{\small
\begin{tabular}{@{}rrrrr@{}}
\toprule
\shortstack[r]{Visual\\positions} &
\shortstack[r]{Prefill\\(ms)} &
\shortstack[r]{Decode\\(ms/token)} &
\shortstack[r]{Visual KV\\(MiB)} &
\shortstack[r]{Peak GPU\\(GiB)} \\
\midrule
144 & 58.8 & 22.9 & 7.9 & 16.44 \\
576 & 143.2 & 22.7 & 31.5 & 16.52 \\
2,304 & 538.2 & 22.8 & 126.0 & 16.86 \\
6,272 & 1414.9 & 23.4 & 343.0 & 17.62 \\
19,600 & 5106.9 & 26.7 & 1,071.9 & 20.20 \\
\bottomrule
\end{tabular}
}
\end{table}

\subsection{Visual Encoding Cost}
\label{app:visual_profile}

We time visual encoding on sixty ActivityNet test videos with 100 sampled frames each. The visual modules run in fp32 and the LLM in bf16. After warm-up on two additional videos, each video is measured three times; we take its median and then average across videos. Individual components and the complete visual encoder are timed separately.

\begin{table}[t]
\centering
\caption{Visual encoding time on sixty videos with 100 sampled frames each. Component times are measured independently, taking the median of three repetitions per video and then averaging across videos.}
\label{tab:supp_visual_timing}
\setlength{\tabcolsep}{5pt}
{\small
\begin{tabular}{@{}lr@{}}
\toprule
Component & Mean time/video (ms) \\
\midrule
SigLIP & 4631.72 \\
Feature adapter & 14.45 \\
Slot attention & 100.35 \\
Slot predictor & 41.32 \\
Observation extractor & 51.49 \\
Trajectory encoder & 11.46 \\
Confidence heads & 0.49 \\
Persistent memory & 318.38 \\
Typed projector & 82.38 \\
Whole visual forward & 5217.07 \\
Scene-cut detection & 156.78 \\
\bottomrule
\end{tabular}
}
\end{table}

SigLIP accounts for most of the visual encoding time (Table~\ref{tab:supp_visual_timing}). The object-state sequence is built once per video and reused for every question about it.

\subsection{End-to-End Inference}
\label{app:system_profile}

We run LLaVA-OneVision-7B in fp16 with PyTorch scaled dot-product attention (SDPA) on the same GPU and videos, at its 32-frame default and at 100 frames to match \ours{}. With 196 pooled patch positions per frame plus one video newline position, these settings give 6,273 and 19,601 visual positions; the context sweep uses the patch-only lengths 6,272 and 19,600.

Table~\ref{tab:supp_system_profile} separates video decoding from model compute. LLaVA-OneVision is timed directly with a fixed 32-token output, including its processor. For \ours{}, the 32-token compute time is composed per video from measured components,
\begin{equation}
 t_{32}=t_{\mathrm{cut}}+t_{\mathrm{visual}}+t_{\mathrm{prefill}}+31t_{\mathrm{decode}},
\end{equation}
with $t_{\mathrm{decode}}\approx0.028$ s/token measured in video inference. Values are medians across videos.

At 100 frames, model compute is 6.29 s for \ours{} and 7.51 s for LLaVA-OneVision; the 32-frame LLaVA-OneVision default takes 2.73 s. Most of the compute of \ours{} is visual encoding, which runs once per video.

\begin{table}[!htbp]
\centering
\caption{Inference cost on the same videos. Model compute excludes video decoding. \ours{} standardizes 32-token compute time from component timings; LLaVA-OneVision generates 32 tokens directly. Peak GPU memory is the median of the per-video maximum across visual encoding and generation.}
\label{tab:supp_system_profile}
\setlength{\tabcolsep}{1.7pt}
{\small
\begin{tabular}{@{}lrrrrr@{}}
\toprule
Execution setting &
\shortstack[r]{Visual\\pos.} &
\shortstack[r]{Video\\decode (s)} &
\shortstack[r]{Prefill\\(s)} &
\shortstack[r]{Compute\\(s)} &
\shortstack[r]{Peak\\(GiB)} \\
\midrule
\shortstack[l]{\ours{}\\100 frames} & 144 & 5.42 & 0.060 & 6.29 & 20.2 \\
\addlinespace
\shortstack[l]{LLaVA-OneVision\\32 frames} & 6,273 & 5.30 & 1.426 & 2.73 & 16.9 \\
\addlinespace
\shortstack[l]{LLaVA-OneVision\\100 frames} & 19,601 & 5.31 & 5.159 & 7.51 & 20.8 \\
\bottomrule
\end{tabular}
}
\end{table}

\end{document}